\documentclass{article}

\usepackage{amsmath}
\usepackage{graphicx}
\usepackage{natbib}
\usepackage{multirow}
\usepackage{adjustbox}
\usepackage{cleveref}
\usepackage{acronym}
\usepackage{csquotes}
\usepackage{url}
\usepackage{eufrak}
\usepackage{authblk}
\usepackage{booktabs}
\usepackage{arydshln}
\newcommand\representation{\mathfrak{R}}
\newcommand\dimension{\mathcal{D}}
\newcommand\relation{\mathcal{R}}
\newcommand\orientation{\mathcal{O}}
\newcommand\feeling{\mathfrak{f}}
\newcommand\change{\mathcal{C}}
\acrodef{ai}[AI]{Artificial Intelligence}
\acrodef{lsc}[LSC]{Lexical Semantic Change}
\acrodef{nlp}[NLP]{Natural Language Processing}
\acrodef{wsd}[WSD]{Word Sense Disambiguation}
\acrodef{lm}[LM]{Language Model}
\acrodef{llm}[LLM]{Large Language Model}
\acrodef{coha}[COHA]{Corpus of Historical American English}

\usepackage{draftwatermark}

\SetWatermarkText{Preprint, please cite \url{https://doi.org/10.1017/nlp.2026.10023}}
\SetWatermarkColor[gray]{0.5}
\SetWatermarkFontSize{1cm}
\SetWatermarkAngle{90}
\SetWatermarkHorCenter{20cm}

\begin{document}
\label{firstpage}

\title{Survey in Characterizing Semantic Change}

\author[1,2]{Jader Martins Camboim de S\'a}
\affil[1]{FSTM - University of Luxembourg\\ 2 place de l’Université, L-4365, Esch-sur-Alzette, Luxembourg}

\author[2]{Marcos Da Silveira}
\author[2]{C\'edric Pruski}
\affil[2]{Luxembourg Institute of Science and Technology,\\ 5 avenue des Hauts-Fourneaux, L-4362, Esch-sur-Alzette, Luxembourg\\
\texttt{first.second@list.lu}}

\maketitle


\begin{abstract}
Living languages continuously evolve to reflect the cultural changes of human societies. This evolution manifests through neologisms (new words) or the \textbf{semantic change} of existing words (new meanings for existing words). Understanding the meaning of words is vital for interpreting texts from different cultures (regionalisms or slang), domains (e.g., technical terms), or time periods. In computer science, this phenomenon is relevant to computational linguistics tasks such as machine translation, information retrieval, and question answering. Semantic change can impact the performance of these applications, making it important to understand and characterize these changes formally.
This problem has recently attracted significant attention from the computational linguistics community. Several approaches can detect semantic changes with good precision, but more effort is needed to characterize \textit{how} word meanings change and to determine how to mitigate the impact of this phenomenon.
This survey provides a comprehensive overview of existing approaches to the \textbf{characterization of semantic change}. We also formally define three classes of characterization: change in dimension (whether a word's meaning becomes broader or narrower), change in orientation (whether a word acquires a more pejorative or ameliorative sense), and change in relation (whether a word is used in a new figurative context, such as a metaphor or metonymy). We demonstrate the applicability of this formalism on existing corpora, summarize the key aspects of selected publications, and discuss current needs and trends in research on semantic change characterization.
\end{abstract}

\maketitle

\section{Introduction}
Language is one of the most dynamic skills humans have acquired, constantly evolving to adapt communication to new purposes and needs \citep[p.~13-23]{pinker2003language}. While no single unified theory explains its origins \citep[p.~13-52]{allan2013oxford}, language has clearly undergone significant modifications throughout human history. One of the prevailing theories of language evolution is based on the assumption of ``\textit{continuity} of change'' \citep{rilling2012continuity}.

This theory posits that language, being too complex to have appeared suddenly, must have evolved gradually from earlier pre-linguistic systems used by our ancestors \citep[p.~725-726]{Pinker1990NaturalLA}. In this context, comprehending language change contributes to the fields of history, sociology, linguistics, philosophy, and \ac{nlp} \citep[p.~1-8]{campbell2013historical}, as it is essential to understanding the origins of language, cultural development, grammatical theories, and the relationship between language and thought \citep[p.~426–454]{Sherwood2008ANH}.

Language change is driven by many unplanned factors, such as linguistic evolution, sociocultural and historical influences, and technological advancements. A variety of grammatical factors also promote changes, including:

\begin{itemize}
    \item \textbf{Phonological Changes:} Changes in the pronunciation and sound systems of a language. For instance, the pronunciation of vowel sounds changed significantly due to the Great Vowel Shift in Medieval English \citep[p.~367]{nevalainen2016oxford}.
    \item \textbf{Orthographic Changes:} Alterations in spelling conventions, often reflecting phonological changes, e.g., Latin \textit{panem} to French \textit{pain} (`bread').
    \item \textbf{Syntactic Changes:} Changes in grammar and word order. For example, Modern English emphasizes word order and auxiliary verbs more than the intricate inflectional system of Old English \citep[p.~616]{nevalainen2016oxford}.
    \item \textbf{Semantic Changes:} Words or phrases acquiring new meanings over time, e.g., \textit{notebook} from ``a book for notes'' to ``a portable computer.''
\end{itemize}

While phonological and morphological changes are crucial, semantics plays a pivotal role in language evolution. Semantic change reflects how words adapt to evolving cultural, technological, and societal contexts, creating new meanings and preserving language's fundamental purpose of effective communication. Despite being a recurring topic in scientific research, semantic change remains less understood than other forms of linguistic evolution due to its often subtle nature compared to more observable shifts, like those resulting from grammar reforms \citep[p.~2-3]{allan2012current}.

Historically, the study of semantic change relied on manual analysis. Linguists carefully examined historical texts, dictionaries, and corpora to track how word meanings evolved. While these studies provided valuable insights, they were slow, labor-intensive, and difficult to replicate.

The advent of computational linguistics and advanced models marked a major shift in the field. With vast digital corpora and efficient algorithms, researchers began to apply statistical methods from natural language processing to analyze text automatically. This marked a transition from purely observational methods to data-driven statistical analysis \citep[p.~161-164]{allan2012current}.

With access to large text corpora and powerful algorithms, linguists can now detect subtle patterns in language with greater accuracy and efficiency. This allows for the study of languages with limited prior linguistic knowledge \citep{Conneau2017WordTW, Wu2019BetoBB}, given sufficient data, and enables comparison of multiple languages to test for universal properties in latent space \citep{Vulic2020AreAG, Wang2022EnglishCL}.

By adopting a model-based approach, linguists can now explore complex language phenomena across different cultures and backgrounds \citep{Nivre2020UniversalDV}. These computational tools provide a structured way to test hypotheses, refine mathematical models, and validate findings against real-world linguistic data \citep{Hammarstrm2019OnCH}.

While initially a topic of theoretical interest, the computational study of semantic change has intrinsic value for understanding language evolution and supporting linguistic research. Beyond its linguistic importance, it also informs certain computer science tasks that involve historical language or evolving terminology. For example, historical knowledge graphs such as Wikidata need to account for changes in entity and relation meanings over time to maintain consistency \citep{Polleres2023HowDK}. In digital humanities, tracking the semantic shift of key concepts (e.g., liberal, gay, computer) over centuries enables more accurate interpretation of historical texts. Similarly, domain-specific information retrieval can benefit from awareness of semantic drift — for example, the term cloud in technical literature has shifted significantly over the past decades.

Developing cost-effective tools that can handle linguistic variation and precisely characterize semantic changes is crucial for many applications. While the phenomena causing semantic change have been studied extensively in Linguistics \citep{traugott2017semantic}, providing deep analyses and typologies, and considerable effort has been devoted to \textit{detecting} such changes by the computational linguistics community \citep{Tahmasebi2018SurveyOC}, to the best of our knowledge, few efforts have focused on formalizing the problem to characterize the \textit{type} of change a word has undergone \citep{Hengchen2021ChallengesFC}.

A formal, mathematical language can serve as a valuable tool in the computational study of semantic change. While we acknowledge that language evolution is deeply intertwined with cultural, societal, and historical dynamics—factors that often resist formalization—developing a structured framework allows us to isolate and investigate specific computational aspects of this complex phenomenon. Our aim is not to reduce semantic change to purely formal terms, but to provide a simplified, tractable representation that supports the development of methods and evaluation tools within computational linguistics. We recognize that such formalizations are necessarily reductive and do not capture the full richness of linguistic and cultural processes; rather, they offer a complementary perspective that can contribute to broader, interdisciplinary understandings of meaning change.

This survey addresses the need to understand and formalize the characterization of semantic change. Characterizing such changes will open new perspectives, particularly since we often lack the resources to study them in detail\footnote{While NLP resources have hundreds of thousands of examples, datasets for semantic change generally cover fewer than a hundred words \cite{Kutuzov2022ContextualizedEF}.}. This is especially problematic in contexts where multiple word meanings are relevant, potentially leading to political or social misinterpretations. This survey's objectives include:
\begin{enumerate}
    \item Identifying the limitations of current representations and methodologies in categorizing semantic changes.
    \item Evaluating the strengths of approaches used to classify \ac{lsc}.
    \item Finding and analyzing formal definitions based on first-order logic to reduce ambiguity and facilitate the comparison of findings.
    \item Demonstrate conceptually the task of \ac{lsc}.
\end{enumerate}

In this survey, we select categories that are both addressable by computational methods in a comparative setting \citep{Tang2013SemanticCC} and well-established in the linguistic literature \citep[p.~221-246]{campbell2013historical}, i.e., changes that can be systematically measured and are backed by theoretical analysis \citep{Hammarstrm2019OnCH}. For instance, euphemistic shifts often depend on social norms. When the word `weed' acquired the sense of `cannabis', this shift was motivated by the intention to obscure the literal meaning, given that cannabis carried negative or prohibited connotations within that particular cultural and historical context. Capturing this type of change requires not only knowledge of the lexical relation between `grass' and `cannabis', but also an understanding of external social and cultural factors.

Another example concerns cases in which a word’s sense becomes more general or more specific. Such changes cannot always be detected through direct comparison of the two senses, as the more general sense may still appear appropriate in both contexts, thus obscuring the shift. We focus on three factors of semantic change: \textbf{orientation} (i.e., whether a word's meaning becomes more pejorative or positive, for example, the word awful acquired a negative meaning), \textbf{relation} (i.e., whether a word's meaning changes through figurative usage, `head' used to address the director of a company), and \textbf{dimension} (i.e., whether a word's meaning becomes broader or more specific, `cell' used to mean only a room in prison).

These poles of change also encode fundamental aspects of meaning: denotative (dimension), connotative (orientation), and cognitive (relation). Some of these properties, as we detail later, are crucial for understanding the intent of a sentence, for writing tools, to assist with translation, or to interact with users (e.g., in a social robot). Finally, this will open new opportunities to improve the reasoning capabilities of \ac{nlp} tools \citep{Danilevsky2020ASO}.

To our knowledge, this is the first survey to comprehensively cover semantic change characterization by consolidating and categorizing existing approaches and identifying research gaps. It introduces formal descriptions and proposes enhancements for a more robust understanding of semantic change. By synthesizing prior research and addressing its limitations, this survey aims to streamline existing work and advance our understanding of this complex phenomenon.

The remainder of the paper is structured as follows: \Cref{sec:background} reviews existing surveys, highlights their differences from this work and present our survey method. In \Cref{sec:corpora}, we investigate available corpora for semantic change analysis. In \Cref{sec:representation}, we explore methods for representing and characterizing meaning change. In \Cref{sec:formalism}, we present a formal model derived from the reviewed methodologies. \Cref{sec:discussion} presents our analysis and open questions. Finally, \Cref{sec:conclusion} concludes the survey and outlines future work.

\section{Lexical Semantic Change}
\label{sec:background}
For the characterization of change, we adopt the mainstream linguistic typology for semantic change \citep{traugott2017semantic}: (i) broadening and narrowing, (ii) amelioration and pejoration, and (iii) metonymization and metaphorization. Before studying change characterization, we first review work on change \textit{identification}. This is a well-developed topic that focuses on evaluating whether a word's meaning has changed. Papers addressing change identification often state the need for further characterization of the type of change, but the state of computational change characterization remains unclear. This conclusion motivated the work presented in this paper and differentiates our survey from existing ones on \ac{lsc}.

In the following subsections, we present existing surveys on the semantic change problem and identify gaps in the current literature. Inspired by their definitions, we propose our own typology of changes. The last subsection introduces the objectives of this survey and our plan to address the gap in semantic change characterization.

\subsection{Related Works on Identification of Lexical Semantic Changes}
Existing surveys on semantic change regularly cover detection but seldom focus on characterization, which is the topic of this paper. In the context of identification, \citet{Tang2018ASO} proposes splitting \ac{lsc} analysis into four steps: obtaining and preparing a diachronic corpus, identifying word senses, modeling semantic change, and presenting and validating hypotheses. This empirical, rather than historical, paradigm enables the study of semantic change through scientific and statistical methods, allowing findings to be reproduced and tested—a key element for computational research.

Tang's framework structures the survey by describing studies related to each step. Change identification is mainly discussed in the ``modeling semantic change'' section, which explains the difference between historiographical and empirical analysis. However, it does not cover approaches that classify changes into categories like broadening, narrowing, pejoration, amelioration, or figurative shifts.

A second survey on computational approaches to \ac{lsc} \citep{Tahmasebi2018SurveyOC} focuses on word-level differentiation, sense differentiation, and lexical replacement. While it includes some types of change categories (e.g., metaphor, metonymy), the main change categories discussed are broadening, narrowing, and correlated types (split, join, birth).

However, one recommendation from the authors is to conduct deeper studies on methods for automatically detecting other types of change and using them to build word histories. The authors provide an in-depth analysis of the linguistic literature grounding each domain, followed by evaluation strategies and applications. The difficulty of comparing approaches is highlighted and attributed to the absence of standard metrics and reference datasets, but also to the inherent challenges of the task itself.

\citet{Kutuzov2018DiachronicWE} survey word representation models and data for semantic change identification and describe a framework for comparing these methodologies. The authors focus on defining consistent terminology and coherent evaluation practices. In this work, they claim there's still a gap in studying sub-classifications of change:
\begin{displayquote}
    ``However, more detailed analysis of the nature of the shift is needed. [...] This problem was to some degree addressed by \citet{Mitra2015AnAA}, but much more work is certainly required to empirically test classification schemes proposed in much of the theoretical work [...]''
\end{displayquote}

For the study of \ac{lsc} and \ac{nlp} in general, there are four main ways to represent word meaning: (i) frequency of co-occurrence, (ii) graphs of associated words or meanings, (iii) topic modeling for a list of meaningful terms, and (iv) word embeddings. Initial studies in the field employed graph- or frequency-based methods. Later approaches benefited from advances in the state of the art, replacing discrete methods with those based on static and contextual embeddings. Although contextual embeddings represented a massive improvement for a series of \ac{nlp} tasks \citep{Devlin2019BERTPO}, the improvement for \ac{lsc} did not follow this trend \citep{Schlechtweg2020Semeval}, showing that there is still room for improvement in this domain \citep{Zhou2021FrequencybasedDI, Kutuzov2022ContextualizedEF}.

Other surveys in \ac{lsc} investigate the phenomenon of change at the ontological level, trying to discover which laws govern the semantic evolution of words. For example, which kinds of words are more prone to broadening, given their current properties? In this context, \citet{Dubossarsky2017OuttaCL} investigate existing evidence for claims such as (i) word frequency affects meaning change, (ii) a negative correlation exists between meaning change and prototypicality, and (iii) a positive correlation exists between meaning change and polysemy. The authors observed that laws proposed in previous studies are affected by model or data bias. When other investigators reduced the bias, the `laws' either did not hold or were less precise than initially reported.

Research in this field also delves into methodological issues, including the challenges of characterizing semantic shifts. As pointed out by \citet{Hengchen2021ChallengesFC}, few studies detail the nature of these changes, mainly because it is difficult to model meaning from textual data. The paper further concludes that no particular measures for quantifying category change exist and that a consensus on a taxonomy for classifying types of change is required.

In the work of \citet{Montanelli2023ASO}, the authors review methodologies based on contextualized word embeddings. They categorize methodologies for training contextualized embeddings based on how they represent meaning, how time information is presented to the model, and what kind of learning procedure is performed. In this paper, the authors also discuss the importance of characterization:

\begin{quote}
    ``Most of the literature papers do not investigate the nature of the detected shifts, meaning that they do not classify the semantic shifts according to the existing linguistic theory (e.g., amelioration, pejoration, broadening, narrowing, metaphorisation, and metonymisation) [...]. These studies could be crucial to detect ``laws'' of semantic shift that describe the condition under which the meanings of words are prone to change.  [...]''
\end{quote}

For this survey, we argue that semantic change characterization needs appropriate methods and tools to analyze each fine-grained category. We also defend that observing variations in a single property cannot describe all forms of \ac{lsc} well. For instance, changes in the angle between word vectors \citep{Dubossarsky2017OuttaCL} is not the best solution to identify the narrowing or broadening of a sense \citep{Tahmasebi2018SurveyOC}. Specific methods for identifying different types of change could lead to a better understanding of \ac{lsc} than straightforward dissimilarity measures.

This survey addresses the need to depict the characterization problem in semantic change and present recent works dealing with this problem. In the next section, we define the survey objectives and the criteria for selecting papers on change characterization.

\subsection{Objectives}
\label{sec:objectives}
Different terms refer to \acf{lsc}, such as semantic drift or semantic shift. In this paper, we distinguish between the identification of a change and the characterization of that change. The former refers to detecting that a word's meaning has changed, while the latter refers to describing how it has changed. We formally differentiate these problems in \Cref{sec:semchange}. The analysis of work on the identification problem has been well covered by \citet{Tahmasebi2018SurveyOC}, \citet{Kutuzov2018DiachronicWE}, and \citet{Tang2018ASO}.

Semantic change occurs from the need to convey more complex concepts and ideas with a restricted set of known words \citep{HockJoseph2019Semantic}. While some authors categorize changes by their cause (e.g., social, cultural), others use linguistic mechanisms (e.g., metaphor, metonymy) as the basis for classification. The focus of our literature review is on the characterization of change, where we look for common properties that allow for grouping identified semantic changes into categories. From a computational perspective, determining the cause of a semantic change is very complex to implement and, to the best of our knowledge, no computational approach has been implemented yet.

Frameworks have been proposed on how to track \ac{lsc}, such as the Cambridge and McTaggart views \citep{Tang2013SemanticCC}, which did not initially consider computational methods. These approaches differ in how they use temporal information to identify changes: while the Cambridge view involves comparing two static corpora from different time periods, the McTaggart view involves tracking the evolution of a word's usage over time. The Cambridge approach best suits the computational perspective, as the McTaggart approach would require historiographical work to acquire socio-cultural knowledge that goes beyond contextual analysis of word occurrences \citep[p.~4-7]{Tang2013SemanticCC}.

This core distinction causes considerable disagreement, as empirical and historiographical methods are based on different ideas and focus on distinct aspects of research. Empirical methods rely on direct observation and measurable data, whereas historiographical methods involve interpreting past events and written records. Since these two approaches operate in different ways and aim to answer distinct kinds of questions, they often yield contrasting views.

For this review, we assume the Cambridge setting for change analysis, using linguistic figures as a basis for change categories, which aligns with current state-of-the-art methods for computational \ac{lsc} \citep{Pivovarova2021RuShiftEvalAS}. Given these considerations, we adopt the typology from \citet{traugott2017semantic}, which is also predominant in the literature, as seen in \citet[p.~21-66]{koch2016meaning}, \citet[p.~189-222]{HockJoseph2019Semantic}, and \citet[p.~221-246]{campbell2013historical}:

\begin{itemize}
    \item \textbf{Broadening:} A word gains new meanings, making it applicable to more concepts, e.g., `cloud' as a computing infrastructure.
    \item \textbf{Narrowing:} A word's meaning becomes more restricted, applying to fewer concepts. For example, `gay' once meant `jolly' or `festive' but now primarily refers to homosexuality.
    \item \textbf{Amelioration:} A word gains a more positive sense. For example, \textit{nice} changed from `foolish, innocent' to `pleasant.'
    \item \textbf{Pejoration:} A word is used with a worse connotation. For example, Old English \textit{stincan} (`to smell sweet or bad') changed to \textit{stink}.
    \item \textbf{Metonymization:} A change based on contiguity or association between concepts, e.g., \textit{board} from `table' to ``a governing body of people who sit at a table.''
    \item \textbf{Metaphorization:} Conceptualizing one thing in terms of another, e.g., using \textit{heart} to represent `feelings.'
\end{itemize}

\begin{figure}[ht]
    \centering
    \includegraphics{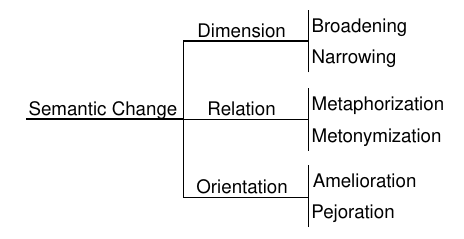}
    \vspace{2em}
    \caption{Taxonomy for the poles of Lexical Semantic Change, based on the work of \citet{koch2016meaning, Blank2003Polysemy, HockJoseph2019Semantic}.}
    \label{fig:taxonomy}
\end{figure}

Since the types in this typology form natural pairs that often involve similar computational challenges (e.g., broadening/narrowing requires sense counting; amelioration/pejoration requires sentiment analysis; metaphorization/metonymization requires a relatedness measure \citep{koch2016meaning, Blank2003Polysemy, HockJoseph2019Semantic}), we group them into three overarching poles to better compare and categorize research. We only include approaches that compare corpora, as this typology is designed for diachronic studies rather than synchronic variation (e.g., across domains). \Cref{fig:taxonomy} illustrates the hierarchical taxonomy for \ac{lsc}.

We assign studies that track sense broadening or narrowing to the \textbf{dimension} ($\dimension$) pole, where words gaining or losing senses see a change in their dimension value (see \Cref{sec:dimension}). This pole is self-complementary, as the change in the dimension's value addresses both broadening and narrowing. A change in dimension could signal, for example, that a word requires more careful translation \citep{Mohammed2009PolysemyAA}.

We group changes related to metaphorization or metonymization into the \textbf{relation} ($\relation$) pole. These changes occur when a new sense develops from an existing one through a figurative link (see \Cref{sec:relation}). Relations like hypernymy, hyponymy, and synonymy are also factors for semantic change \citep{koch2016meaning, HockJoseph2019Semantic} as they involve a conceptual relation and can potentially be encompassed in this pole.

For the \textbf{orientation} ($\orientation$) pole, we group studies that identify amelioration, pejoration, or any perceptual change in the sentiment associated with a word (e.g., criticism, compliment). This pole also encompasses many forms of change in perception, as some studies suggest that concepts like `taboo' also affect changes in word usage \citep{koch2016meaning}.

\begin{figure}
    \centering
    \includegraphics[width=\textwidth]{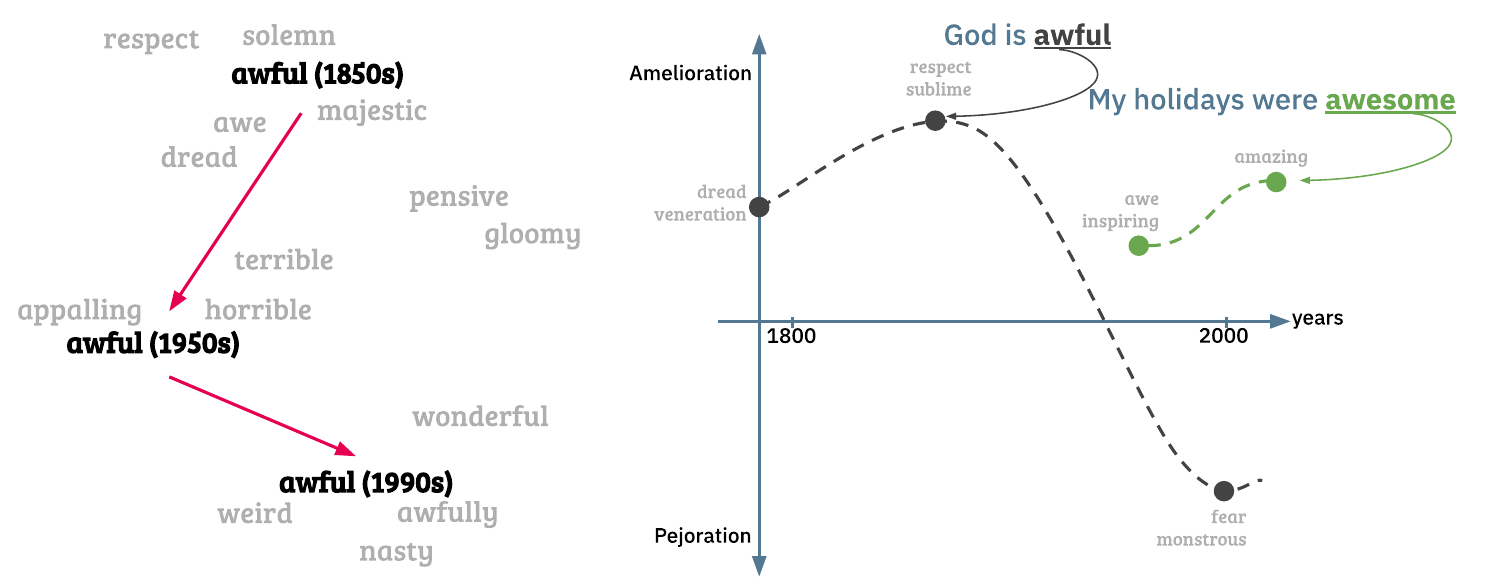}
    \vspace{2em}
    \caption{Change in meaning and orientation for the word awful. In the left side, we reproduce the figure from \citet{Hamilton2016DiachronicWE} that shows the evolution of the word `awful' in the embedding space. In the right side, we present the hypothetical function ($\feeling$) for this word over time.}
    \label{fig:orientation}
\end{figure}

To illustrate a word that changed in orientation, consider `awful'. In the 1850s, it meant ``full of awe,'' but today it means ``extremely disagreeable.'' As the first sense is semantically positive and the second is negative, this word underwent pejoration, as illustrated in \Cref{fig:orientation}. Notice that the word's orientation may have first undergone amelioration before pejoration. In the same figure, the word `awesome,' which has a close relation to "awe," gained a more positive sense, undergoing amelioration over time.

In the next section, we describe our method for reviewing the literature based on the poles presented above.

\subsection{Review Criteria}
In this survey, we select papers that represent and analyze words from one of the poles mentioned above in an automatic manner. We exclude works that involve manual analysis, as we focus on computational methods. For the relation and orientation poles, these searches returned many unrelated results, such as studies on synchronic metaphor detection, which do not infer diachronic change and were therefore excluded.

Querying scientific databases required an iterative process to include all relevant terms. In our initial efforts, we searched for ``[type of change, e.g., broadening] semantic change'' in Scopus, which returned 33 articles; however, none of them were relevant. Due to the lack of consensus in terminology and the rapid pace of the field, we adopted a semantic search approach, starting with key papers and iteratively expanding our search query. \Cref{tab:terms} presents the final set of search terms used to collect articles.

\begin{table}[ht]
    \centering
    \caption{Search terms used to discover articles related to change characterization.}
    {
    \begin{tabular}{cc}
    \hline
         Characterization & Search terms \\\hline
         Dimension & \begin{tabular}{@{}c} ``semantic change broadening'',``semantic change narrowing'',\\``novel sense change'',``innovative meaning'',``new sense identification'',\\``neologism identification''\end{tabular}  \\\hdashline
         Relation & \begin{tabular}{@{}c} ``metonymy semantic change'',``metonymization semantic change'',\\``metaphor semantic change'',``metaphorization semantic change''\end{tabular}\\\hdashline
         Orientation & \begin{tabular}{@{}c} ``semantic change polarity'',``semantic change feeling'',\\``semantic change orientation'',``semantic change emotions'',\\``semantic change sentiment''\end{tabular} \\\hline    
    \end{tabular}} 
    \label{tab:terms}
\end{table}

\begin{figure}[ht]
    \centering
    \includegraphics[width=0.5\textwidth]{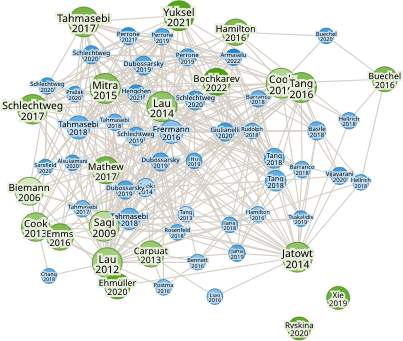}
    \vspace{2em}
    \caption{Graph of selected works (green) and related articles (blue).}
    \label{fig:articles}
\end{figure}

From these initial queries we could find 11 seed papers that satisfy our criteria. We then employed a network semantic citation graph to recover more papers, Research Rabbit\footnote{\url{https://researchrabbitapp.com/}}. From the 11 initial papers, the graph suggested 83 new papers, from this we were able to recover 7 that matched our criteria. We again included these additional 7 papers as seed, obtaining 129 papers. From 129 papers we obtained 5 more relevant papers. In the last interaction we obtained 151 paper suggested but no new relevant papers were found. In \Cref{fig:articles}, we present the graph of documents, where green nodes represent the input papers, and blue nodes are papers suggested by the tool.

In the next sections, we investigate three elements of semantic change characterization in depth: (i) the corpora ($T$) analyzed, (ii) the representation method ($\representation$) for capturing characteristics of change, and (iii) the kind of characterization ($\{\dimension,\relation,\orientation\}$) addressed.

\section{Corpora for Semantic Change Analysis}
\label{sec:corpora}
Corpora for semantic change can be selected based on differences in time (diachronic studies), domain (academic fields \citep{Schlechtweg2019AWO}), culture (social media \citep{Carpuat2013SenseSpottingNL, Shoemark2019RoomTG}, countries), or dialect \citep{Takamura2017AnalyzingSC, Uban2021CrosslingualLO}, provided the corpora share a sufficient vocabulary for comparison.

In diachronic studies, some methods divide a corpus into discrete time windows, striking a balance between the size of the window and the quantity of data it contains. In contrast, others use a continuous representation of time, providing the timestamp of each text directly to the model \citep{Hu2019DiachronicSM, Rosenfeld2018DeepNM, Wang2021Word2FunMW}. These corpora were designed for specific purposes. For example, corpora with discrete time bins enable a comparative perspective on meaning, while continuous time representations allow for an evolutionary view of the changes.

With respect to characterization, the collected papers show an increasing interest in diachronic corpora, with many proposing handcrafted rules for automatically annotating change categories \citep{Tang2013SemanticCC}. Few works rely on human annotation \citep{Schlechtweg2017GermanIF}. While these approaches have advantages and drawbacks, the data required to train large representation models is often insufficient from a computational perspective.

Most of the works reviewed here follow a specific formula for diachronic studies, one historical corpus ($C_1$) and a more modern corpus ($C_2$). Where splitting \ac{coha} \cite{Davies2012ExpandingHI}, a diversified corpus with a span from 1810s to the 2000s, in two corpus is the standard choice. Recent state-of-the-art \ac{nlp} models require training on billions or even trillions of tokens \citep{He2021DeBERTaV3ID}, but diachronic corpora are often on the scale of millions of tokens \citep{Schlechtweg2020Semeval}. Given that increasing the amount of historical training data is often infeasible, a viable approach is to develop better algorithms for learning from limited data.
In \Cref{tab:corpora_overview} we present the table of diachronic corpus in the reviewed works.

\begin{table}[htpb!]
    \centering
    \caption{Overview of Various Corpora for Diachronic Studies}
    \label{tab:corpora_overview}
    \begin{adjustbox}{width=\textwidth}
    \begin{tabular}{p{0.3\textwidth} p{0.4\textwidth} p{0.3\textwidth}}
    \toprule
    \textbf{Corpus Name} & \textbf{Description} & \textbf{Statistics (Date, Size)} \\
    \midrule
    \citeauthor{CLMET} &
    A collection of diachronic English corpora covering various text types like literary works, letters treatise and others. &
    34 million words, spanning from 1710-1920, divided into three 70-year sub-periods.  \\
    \midrule
    \citeauthor{COHA} &
    A balanced diachronic corpus of American English compiled from diverse document genres. &
    Over 400 million words from about 107,000 documents published from the 1810s to the 2000s. \\
    \midrule
    \citeauthor{PDN}&
    A Chinese newspaper corpus used for its faithful recording of contemporary language and events. &
    Spans from 1946 to 2004, with a yearly average of 10,886,017 tokens. \\
    \midrule
    \citeauthor{GoogleBook} &
    The largest available historical corpus, compiled from about 4\% of all ever-published books. &
    Over 1TB of text data, containing about 0.3 trillion words and spanning from 1600 to 2009. \\
    \midrule
    \citeauthor{Kubhist2} &
    A sample of a Swedish historical newspaper archive. &
    From 1790-1830 containing about 71 million tokens. \\
    \midrule
    \citeauthor{DTA} &
    A high-quality collection of historical German texts selected for representativeness and genre balance. &
    1,022 texts from the period 1741-1900. \\
    \midrule
    \citeauthor{LC}  &
    A collection of British English texts, encompassing various domains such as religion, politics, and law. &
    Approximately one million words from 1640-1740. \\
    \midrule
    \citeauthor{BNC} &
    A collection of primarily written British English sources. &
    One hundred million words, from the late 20th century. \\
    \midrule
    \citeauthor{NYT} &
    A collection of newspaper articles from The New York Times archives. &
    13 million articles (1851-present). \\
    \midrule
    \citeauthor{Helsinki} &
    Comprises texts spanning Old English, Middle English, and Early Modern English. The analysis focused on the Middle and Modern English. &
    Approximately 1.1 million words. Old (\~1150), Middle (1150-1500), and Modern (1500-1710)\\
    \midrule
    \citeauthor{Gutenberg} &
    The bulk of English language literary works available through the project's website. &
    4034 separate documents consisting of over 290 million words. \\
    \bottomrule
    \end{tabular}
    \end{adjustbox}
\end{table}

Even with progress in low-resource learning fields like rapid adaptation \citep{Neubig2018RapidAO}, few-shot learning \citep{Liu2022FewShotPF}, or in-context learning \citep{Min2022RethinkingTR}, there is still a significant lack of freely available corpora for semantic change studies \citep{Kutuzov2018DiachronicWE}.

\section{Methods for Semantic Change Characterization}
\label{sec:representation}
While interest in detecting semantic change is growing, less effort is being applied to characterizing these changes. Computational linguistics has not yet fully explored the utility of having detailed information about the type of change. For example, a change in relation might have a low impact on a sentiment analysis algorithm, whereas a change in orientation is more likely to affect it.

In this context, meaning representation ($\representation$) methods are highly relevant to understanding what kind of change can be described and characterized. This section investigates representation methods from the literature based on word frequency, topics, graphs, and embeddings. We analyze how each method was used to characterize \ac{lsc} within our three proposed poles.

\subsection{Word frequency}
Methods based on word frequency were among the first computational attempts to tackle semantic change and remain central to many linguistic approaches. Analyzing word frequency offers a direct interpretation of change, which, for characterization, usually involves counting co-occurrences with other words.

\subsubsection{Dimension}
To detect changes along the dimension axis ($\dimension$), \citet{Sagi2009SemanticDA} proposed counting co-occurrences of words and phonemes and measuring the cluster density of the cosine distance on a decomposition of the term-frequency matrix. They assumed that Latent Semantic Analysis is appropriate for examining vectors representing the context of each term's occurrence and estimated cohesiveness by measuring the density of these context vectors. While they explored the methodology for taxonomic change analysis, they state that the method in fact measures word polysemy.

\citet{Tang2013SemanticCC} proposed a framework for semantic change computation based on the assumption that it is an inherently successive and diachronic process. They distinguished between the word level (overall meaning) and the sense level (individual senses), building their approach around these two notions. A key step involves first identifying a word's senses in a corpus and then analyzing their diachronic distribution.

\citet{Emms2016DynamicGM} proposed a generative conditional model to detect if a word has acquired a new sense (a semantic neologism). They modeled the word’s sense distribution using Gibbs sampling, applying the model to time-stamped text to infer how the probabilities of a word's senses change over time. The authors obtain all usages for a word in a specific year ($C_1$) from Google N-gram Corpus and compare with the usage frequency for the following next year($C_2$), obtaining n-grams that change above a threshold.

\citet{Bochkarev2022NeuralNA} used a neural network to predict whether a word is used as a named entity. With this approach, they constructed a temporal view of word usage by tracking occurrences as nouns versus non-nouns over time, aiming to detect if the word acquired a new meaning.

\subsubsection{Relation}
The work of \citet{Tang2013SemanticCC} transforms the problem of relation identification into detecting change patterns in time series data. They analyze semantic change at the word and individual-sense levels, using rules to infer whether a word is used metonymically or metaphorically. A different approach computes the entropy of target words to identify if a word was used in a non-standard context \citep{Schlechtweg2017GermanIF}. The authors split the usage of words from the Deutsches Textarchiv (DTA) in two time periods (varying by word) and measure the increase in entropy as a proxy for metaphorization. According to the authors, a word's higher entropy makes it more likely to be used in a newly created metaphorical sense. However, the authors do not differentiate this from the emergence of a new, unrelated sense. They use a corpus of German documents to validate their approach.

\subsubsection{Orientation}
Frequency metrics that detect co-occurrences with sets of ``good'' or ``bad'' sentiment words have been widely used to infer the amelioration and pejoration of a word over time, especially for neologisms \citep{Cook2010AutomaticallyIC, Jatowt2014AFF}. Similarly, \citet{Buechel2016FeelingsFT} used seed words in German and English to track emotions captured in corpora. They represent these emotions in the Valence-Arousal-Dominance framework rather than simple positive/negative connotation. A limitation of the latter study is that it doesn't track the evolution of orientation in words.

\subsection{Topics}
Topic modeling is a statistical method used to identify the main themes in a collection of documents by grouping words that frequently appear together. For example, a topic model might identify the topic of ``banking'' in a news corpus by grouping words like ``government,'' ``money,'' and ``insurance.''

Topic modeling can be used to understand the meaning of words and phrases by tracking the different topics a word is associated with in various contexts (e.g., time, domain, culture). For example, a topic model could be used to understand the different meanings of ``bank'' in a financial context versus a geographical one. The different topics associated with a word are tracked and compared to identify and characterize changes in its senses.

\subsubsection{Dimension}
These approaches assume that changes in the number of topics assigned to a word can indicate changes in dimension. \citet{Lau2012WordSI} introduced the idea of novelty as the proportion of a word's usages corresponding to senses in a focus corpus versus a reference corpus. The goal is to use word sense induction based on topic modeling to detect if a word is used with a new meaning. 

\citet{Cook2013ALA} extended this work by quantifying a topic's usage, estimating that a previously "culturally silent" topic that becomes frequent in a focus corpus is highly likely to contain neologisms. They introduced the concept of topic relevance to compute a ranked list of potential neologisms.

\begin{figure}
    \centering
    \includegraphics[width=\linewidth]{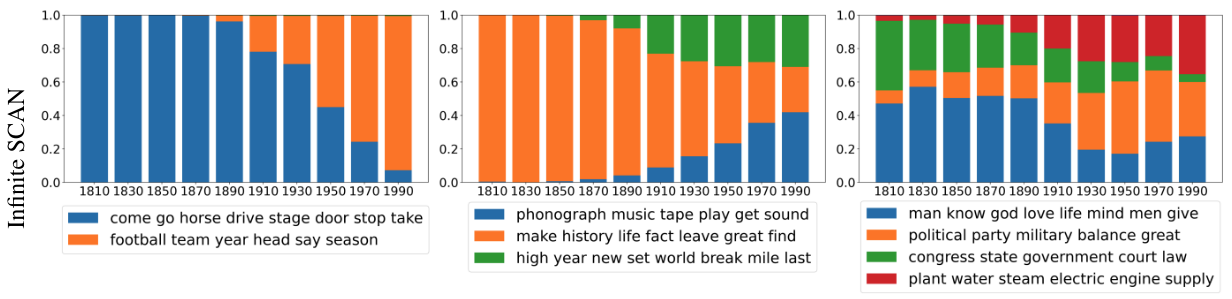}
    \vspace{1em}
    \caption{Figure adapted from \cite{Inoue2022InfiniteSA}. The stacked bar plots represent the topics obtained over time for the words `coach', `record' and `power' respectively. We can observe new senses emerging and becoming dominant.}
    \label{fig:infinitescan}
\end{figure}

\citet{Lau2014LearningWS} improved the definition of novelty by adding the word's frequency in the corpus, then proposed combining novelty and relevance to rank neologisms. \cite{Frermann2016ABM} introduced SCAN, a Bayesian topic model where topics are preserved across adjacent time periods. \citet{Inoue2022InfiniteSA} introduced InfiniteSCAN, an improvement over SCAN. While most topic models rely on a predefined number of topics, InfiniteSCAN uses a logistic stick-breaking process to adapt the number of topics dynamically. This modification allows it to track a varying number of senses, enabling to count each meaning to measure broadening and narrowing, as shown in \Cref{fig:infinitescan}. The authors split Clean \ac{coha} in intervals of 20 years to represent the senses.

\subsubsection{Relation}
To the best of our knowledge, no methodology for tackling conceptual change with topic-based representation has been published, although a work uses it in a synchronic detection setting \citep{Heintz2013AutomaticEO}.

\subsubsection{Orientation}
To the best of our knowledge, no methodology approaching orientation change with topic-based representation has been published. \cite{Wankhade2022ASO} use topic models in their workflow to refine data for a sentiment classifier, but topics are not directly used to detect change in feelings.

\subsection{Graphs}
Graph structures are a natural way to represent relationships between words. They are often constructed manually from dictionaries or automatically based on word co-occurrence.

Co-occurrence graphs can capture richer contextual information than count-based methods because their edges represent more complex relationships. For example, the words `feline,' `cat,' and `pet' would likely form a clique in a co-occurrence graph, indicating they are often used together and share close meanings.

\subsubsection{Dimension}
Comparing the size and structure of a graph over time can provide rich information about a word's semantic change. \citet{biemann2006chinese} was among the first to use this approach to detect sense broadening. He created a graph of associated nouns from a corpus and then applied the Chinese whispers algorithm to find clusters. By comparing these clusters over time, he could infer whether a word had acquired a new meaning.

\citet{Mitra2015AnAA} analyzed tweets, counting word co-occurrences in bi-grams to build a graph. They ran a graph-clustering algorithm (Chinese Whispers) to determine a set of senses, where each cluster represents a different sense. By comparing these sets of senses over time, they could infer if a word's senses split or became broader.

\begin{figure}
    \centering
    \includegraphics[width=0.7\linewidth]{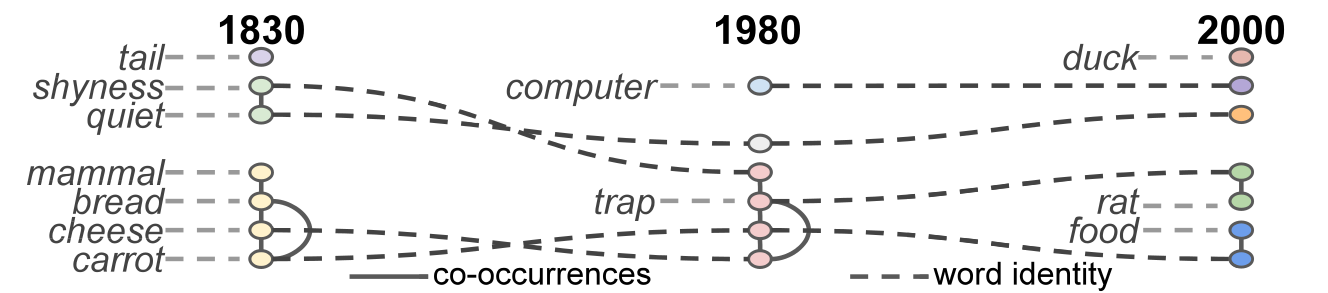}
    \vspace{3em}
    \caption{Adaptation from \citet{Ehmller2020SenseTD}. Ego-network, built from word co-occurrence graph, for `mouse'. We observe that in 1830 it was used with the sense of `weak' and `rat,' where in 1960 the sense of `computer device' emerged.}
    \label{fig:sensetree}
\end{figure}

\citet{Mathew2017Adapting} improved the filtering and sense selection methods of previous works \citep{Mitra2015AnAA, Mccarthy2004Finding, Lau2014LearningWS} to better capture sense evolution in Google Books and newspaper datasets.
\citet{Tahmasebi2017FindingIW} considered a graph of two-word noun phrases as a cluster of meaning, then merged sub-graphs based on Lin similarity \citep{Lin1998AnID}. They tracked change in dimension by analyzing how these sub-graphs change over time.

\citet{Ehmller2020SenseTD} built a co-occurrence graph, filtering it with point-wise mutual information to remove non-relevant edges, and then created ego-networks to run graph-clustering algorithms. This approach allowed them to build a sense tree—a tree of word usage evolution where new senses are new branches, as shown in \Cref{fig:sensetree}. Similar to previous work, the authors restricted their analysis to nouns only.

\subsubsection{Relation}
Graphs are a good way to represent word relations, and some approaches have used them to detect metonymy usage, e.g., \cite{Teraoka2016MetonymyAU, Pedinotti2020Dont}. Their models use WordNet or a graph-based distributional model to determine a contextualized representation of word semantics. However, to the best of our knowledge, no work has proposed using graphs at a representation level to compare which words have changed their relations between corpora.

\subsubsection{Orientation}
\begin{figure}
    \centering
    \includegraphics[width=0.8\linewidth]{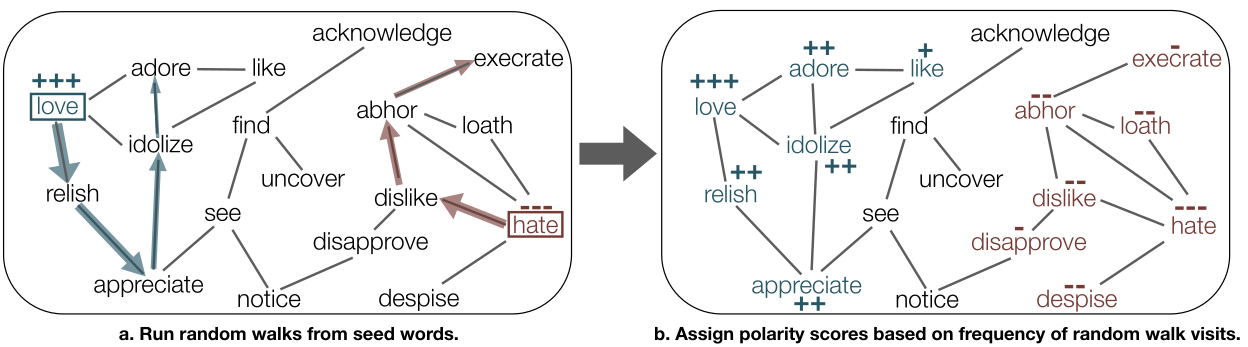}
    \vspace{2em}
    \caption{Adapted from \citet{Hamilton2016InducingDS}. The SentiProp algorithm propagates the polarity from seed words based on distance of the connected nodes. Words are connected based on co-occurence statistics.}
    \label{fig:sentprop}
\end{figure}

The orientation of words is generally inferred from similarity or proximity to seed words. Many word graphs are constructed for this purpose, and some works \citep{Reagan2017SentimentAM, Barnes2021StructuredSA} use these graphs to infer orientation. \citet{Hamilton2016InducingDS} used the SentiProp algorithm to propagate a word's polarity to its synonyms and connected words, analyzing changes in polarity for words across different times and domains. In \Cref{fig:sentprop}, we reproduce the illustration of this algorithm.

\subsection{Embeddings}
Word embeddings are representations of words as numeric vectors. The vectors are constructed such that words with close values in the vector space are expected to have similar meanings. The distance between these vectors can help in interpreting their sense.

Word embeddings have been used for various tasks, including \ac{lsc} identification and characterization. This is done by comparing word vectors between two moments in time; the distance between them, and relative to neighboring vectors, can indicate if the word's sense has changed \citep{Hu2019DiachronicSM}.

\subsubsection{Dimension}
\citet{Ryskina2020WhereNW} proposed a statistical approach to identify neologisms based on frequency count and semantic sparsity (measured with embeddings). They claim that words and concepts are subject to supply and demand, so emerging concepts (i.e., neologisms or sense broadening) appear to populate sparse semantic regions until the space becomes uniformly covered.

In \cite{Giulianelli2020AnalysingLS}, the authors use BERT to create contextualized embeddings of words and then cluster them to obtain usage types. They apply entropy difference, Jensen-Shannon divergence (JSD), and average pairwise distance (APD) as proxies to measure word broadening and narrowing, finding that higher APD and JSD were the best metrics for broadening.

\begin{figure}
    \centering
    \includegraphics[width=0.5\linewidth]{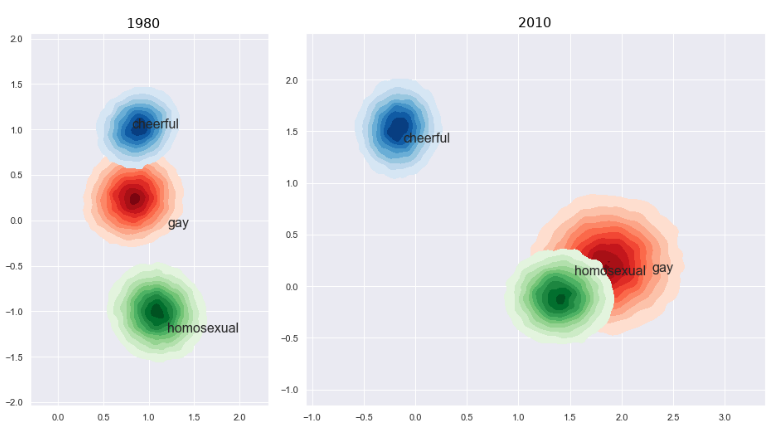}
    \vspace{2em}
    \caption{Plot obtained from \citet{moss2020detecting}. The author represent words as gaussian embeddings, and analyze it's variance and proximity over time. The word gay increased in variance and got closer to `homosexual.'}
    \label{fig:gaussianemb}
\end{figure}

By analyzing the deviation of a Gaussian word embedding, \citet{Yksel2021SemanticCD} identified a way to measure narrowing and broadening in word senses. In \citet{moss2020detecting}, the authors also performed per-sense training of these embeddings. In \Cref{fig:gaussianemb}, we observe the distribution of the word `gay' becoming wider and closer to `homosexual'.

\cite{Lautenschlager2024DetectionON} uses XL-LEXEME \citep{Cassotti2023XLLEXEMEWP} clustered embeddings to progressively discover new senses in the historical COHA corpus. They frame the task as Unknown Sense Detection: given an initial dictionary, the goal is to identify unrecorded senses. The system compares a word usage to all its dictionary senses, and if the highest similarity score falls below a threshold, the usage is flagged as a potential new sense.

\subsubsection{Relation}
\citet{Fonteyn2021AdjustingSA} investigate the metaphorization of the concept ``to death.'' They combined cluster analysis and sentiment analysis to contextualize occurrences of this phrasal verb and detect if metaphorization had taken place. Another approach is to create a training corpus and use it to train a BERT model to predict if a word is metaphorical or literal. The work of \citet{Maudslay2022MetaphoricalPD} follows this approach and proposes a framework to detect words that started being used metaphorically.

\subsubsection{Orientation}
\begin{figure}
    \centering
    \includegraphics[width=0.8\linewidth]{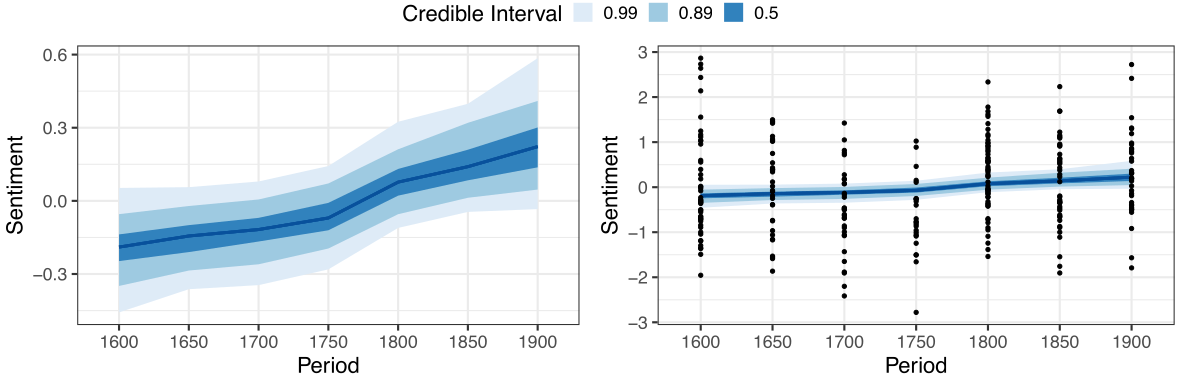}
    \vspace{2em}
    \caption{Adaptation from \citet{Fonteyn2021AdjustingSA}. The line plot shows the evolution of polarity of the multi-word `to death.' It went from a negative concept to a more positive one, ameliorating the dominant sense.}
    \label{fig:adjscope}
\end{figure}

Methods based on word embeddings have been widely used for many sentiment analysis tasks \citep{Raffel2019ExploringTL} and to investigate sentiment at the lexicon level \citep{Ito2020WordLevelCS}. In the context of \ac{lsc} characterization, \citet{Fonteyn2021AdjustingSA} used the distance between the word vectors of `good' and `bad' to measure the polarity of the expression `to death,' as illustrated in \Cref{fig:adjscope}. 

\begin{figure}
    \centering
    \includegraphics[width=0.7\linewidth]{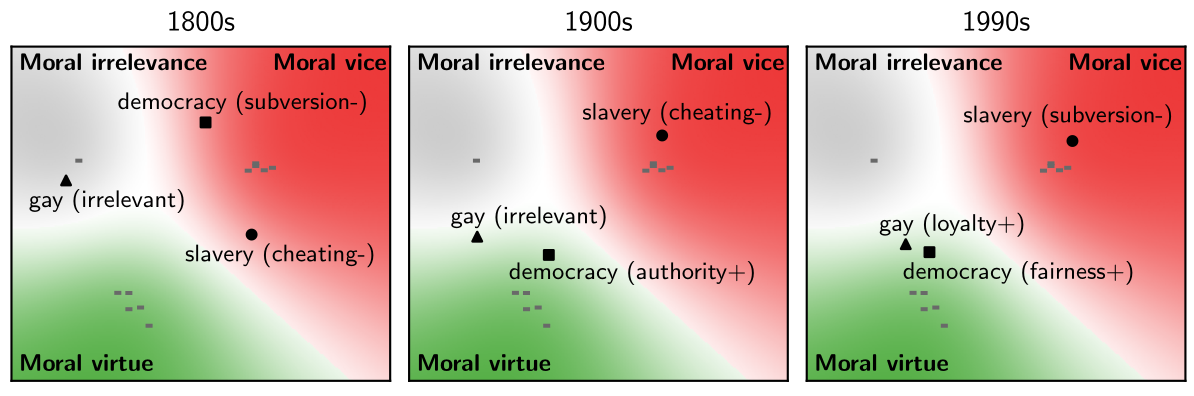}
    \vspace{3em}
    \caption{Illutration adapted from \citet{Xie2019TextbasedIO}. The figure shows how moral sentiment toward slavery, democracy, and gay evolved over two centuries, mapped in a 2D embedding space.}
    \label{fig:moral}
\end{figure}

Similarly, \citet{Hellrich2018ModelingWE} \citet{Xie2019TextbasedIO} employed diachronic static embeddings to analyze the sentiment of words over time. 
Hellrich \textit{et al.} train word embeddings in \ac{coha} from 1810-1830 as $C_1$ and a VAD lexicon \citet{Warriner2013NormsOV} as $\representation (C_2)$, tracking for orientation of words the valence-arousal-dominance framework. They access the polarity of the word based on the k-nearest-neighbors from a VAD sentiment lexicon. Xie \textit{et al.} use `moral' seed words to compare, like `cheating' and 'fairness,' also for \ac{coha} as shown in \Cref{fig:moral}.

\subsection{Synthesis and Summary}
In this section, we synthesize the pros and cons of representation methods for characterizing semantic change and present a table that provides an overview of the current poles of change addressed in the literature.

\textbf{Word frequency}-based methods are simple to implement, offer an easily interpretable analysis, and require a relatively small corpus of data. However, these methods struggle with capturing word similarity \citep{Mikolov2013EfficientEO} and compositionality \citep{Mitchell2010CompositionID}, limiting their semantic analysis capabilities. While N-grams slightly improve compositionality, they are constrained by Zipf’s law \citep{Zipf1949HumanBA}, requiring an exponential increase in data to increase the size of N. Additionally, frequency alone cannot distinguish between sense broadening and narrowing, requiring extra steps to handle polysemy. Co-occurrence tables also fail to correlate well with polysemy or semantic shifts. In parallel studies, these methods perform poorly in detecting word orientation \citep{Socher2013RecursiveDM} and struggle with contexts involving sarcasm, irony, or negation due to their limited contextual awareness. 
\textbf{Topic modeling} is a powerful text analysis tool with many good libraries implementing state-of-the-art algorithms, and it doesn't require supervision or data annotation. The downside is that these methods require manual interpretation, as statistically derived word clusters can be ambiguous. Automated topic selection using dictionary-based filtering aids in neologism detection but remains insufficient for change characterization. Metrics like coherence and diversity, along with visualization techniques, help reduce manual effort.

\textbf{Graph}-based approaches rely heavily on annotated and structured data, which is often costly or, in some domains, unfeasible. However, \acp{llm} enable automatic annotation and structuring of information. We have yet to fully leverage modern structures like knowledge graphs for \ac{lsc} detection. Resources like Wikidata could help model semantic shifts \citep{Giulianelli2023InterpretableWS, Tang2023CanWS}, such as metonymization. Existing lexical databases (WordNet, BabelNet, ConceptNet) assist in change detection but lack temporal information, limiting their effectiveness.

\textbf{Word embeddings} often exhibit instability, where minor corpus variations alter vector representations, especially for low-frequency or polysemous words \citep{Hellrich2016Bad, Gladkova2016Intrinsic, Gnther2019Vector}. Solutions such as larger embeddings, quantization, and alignment help mitigate this instability \citep{Hamilton2016DiachronicWE}; however, the meaning conflation deficiency \citep{CamachoCollados2018FromWT} remains an issue, as a single vector---even when contextualized---does not provide enough information to differentiate fine-grained senses \citep{Ballout2024FOOLMI, Periti2024AnalyzingSC}. Subsequent work has explored various modifications to transformer architectures to enable better representation of meaning \citep{Zhang2021DisentanglingRO, Tay2022UL2UL}, but these approaches remain under-explored in the semantic change literature.

Contextualized embeddings (e.g., from ELMo, BERT) enhance sense detection but can produce overly contextualized vectors, resulting in unreliable diachronic analysis and false positives \citep{Kutuzov2022ContextualizedEF}. Recent methods using contrastive learning and similarity-based training enhance isotropy and sense differentiation \citep{Gao2021SimCSESC, Reimers2019SentenceBERTSE}. Other approaches, like density-based embeddings (e.g., word2gauss), aim to model polysemy directly. The current state of the art in \ac{lsc} identification fine-tunes cross-lingual models trained on \ac{wsd} tasks, benefiting from transfer learning across languages.

While \textbf{generative \ac{ai}}, particularly \acp{llm}, is rapidly transforming NLP, its application to the systematic characterization of semantic change in diachronic data remains largely unexplored. However, some initial studies have successfully leveraged these models for related tasks, such as generating synthetic data for \ac{lsc} detection \citep{Cassotti2024Using} and automatically comparing word senses in context \citep{Sa2024SemanticCC}.

\begin{table}[ht]
\centering
  \caption{Comparison table between studies highlighting the type of methods utilized and category of change. We mark which kind of characterization the work conducts $\dimension$ for dimension, $\relation$ for relation, and $\orientation$ for orientation. Also, we highlight the main representation method ($\representation$) for the word meaning.}
 {
    \begin{tabular}{lcccc}
    \hline
    Methodology & $\dimension$ & $\relation$ & $\orientation$ & $\representation$\\\hline
        \citet{biemann2006chinese} & X & - & - & Graph\\\hdashline
        \citet{Sagi2009SemanticDA} & X & - & - & Frequency\\\hdashline
        \citet{Lau2012WordSI} & X & - & - & Topics\\\hdashline
        \citet{Cook2013ALA} & X & - & - & Topics\\\hdashline
        \citet{Tang2013SemanticCC} & X & X & - & Frequency\\\hdashline
        \citet{Lau2014LearningWS} & X & - & - & Topics\\\hdashline
        \citet{Mitra2015AnAA} & X & - & - & Graph\\\hdashline
        \citet{Emms2016DynamicGM} & X & - & - & Frequency\\\hdashline
        \citet{Mathew2017Adapting} & X & - & - & Graph and Topics\\\hdashline
        \citet{Tahmasebi2017FindingIW} & X & - & - & Graph\\\hdashline
        \citet{Ryskina2020WhereNW} & X & - & - & Embeddings\\\hdashline
        \citet{Giulianelli2020AnalysingLS} & X & - & - & Embeddings\\\hdashline
        \citet{moss2020detecting} & X & - & - & Embeddings\\\hdashline
        \citet{Ehmller2020SenseTD} & X & - & - & Graph\\\hdashline
        \citet{Yksel2021SemanticCD} & X & - & - & Embeddings\\\hdashline
        \citet{Inoue2022InfiniteSA} & X & - & - & Topics\\\hdashline
        \citet{Bochkarev2022NeuralNA} & X & - & - & Frequency\\\hdashline
        \citet{Lautenschlager2024DetectionON} & X & - & - & Embeddings\\\hdashline
        \citet{Schlechtweg2017GermanIF} & - & X & - & Frequency\\\hdashline
        \citet{Fonteyn2021AdjustingSA} & - & X & X & Embeddings\\\hdashline
        \citet{Maudslay2022MetaphoricalPD} & - & X & - & Embeddings\\\hdashline
        \citet{Cook2010AutomaticallyIC} & - & - & X & Frequency\\\hdashline
        \citet{Jatowt2014AFF} & - & - & X & Frequency\\\hdashline
        \citet{Buechel2016FeelingsFT} & - & - & X & Frequency \\ \hdashline
        \citet{Hamilton2016InducingDS} & - & - & X & Graph \\ \hdashline
        \citet{Hellrich2018ModelingWE} & - & - & X & Embeddings\\\hdashline
        \citet{Xie2019TextbasedIO}  & - & - & X & Embeddings \\
    \hline
    \end{tabular}}
  \label{tab:comparison}
\end{table}

\Cref{tab:comparison} summarizes the literature surveyed in the previous sections. The rows list relevant papers, and the columns represent the key properties of \ac{lsc} evaluated in this survey. An ``X'' indicates that the paper applied the sense computation technology (listed in the fifth column) to analyze the corresponding pole.

We can observe that no single methodology characterizes all three poles, highlighting the complexity of the problem and researchers' tendency to focus on a specific type of change. The majority of studies were developed to characterize changes in the dimension pole, which seems to be where computational approaches achieve the best results. Another observation is the under-explored use of graph- and topic-based methods to represent sense, compared to embedding- and frequency-based methods. It is also important to note that the most recent works use embeddings, showing a current trend toward this type of approach. In \Cref{tab:pros_cons} we summarize these findings.

\begin{table}[ht]
\centering
\caption{Pros and Cons of Representation Methods for Semantic Change Characterization}
\label{tab:pros_cons}
\resizebox{\textwidth}{!}{%
\begin{tabular}{@{}llll@{}}
\toprule
\textbf{Pole of Change} & \textbf{Representation} & \textbf{Pros} & \textbf{Cons} \\ 
\midrule

\multirow{8}{*}{\begin{tabular}[c]{@{}l@{}}\textbf{Dimension}\\(Broadening/\\Narrowing)\end{tabular}} 
& \textbf{Frequency} & 
\begin{tabular}[t]{@{}l@{}}- Simple and interpretable. \\ - Good for detecting neologisms (new words).\end{tabular} & 
\begin{tabular}[t]{@{}l@{}}- Struggles to distinguish distinct senses (polysemy). \\ - Can't differentiate sense gain from higher usage (trend).\end{tabular} \\ 
\cmidrule(l){2-4}
& \textbf{Topics} & 
\begin{tabular}[t]{@{}l@{}}- Effective at discovering new senses (sememes). \\ - Unsupervised methods don't require labels.\end{tabular} & 
\begin{tabular}[t]{@{}l@{}}- Requires manual interpretation to label topics (senses). \\ - Topic granularity can be hard to control.\end{tabular} \\ 
\cmidrule(l){2-4}
& \textbf{Graphs} & 
\begin{tabular}[t]{@{}l@{}}- Intuitive for modeling sense splitting and merging. \\ - Can create visual, interpretable sense histories.\end{tabular} & 
\begin{tabular}[t]{@{}l@{}}- Performance is sensitive to graph construction methods. \\ - Can be computationally intensive.\end{tabular} \\ 
\cmidrule(l){2-4}
& \textbf{Embeddings} & 
\begin{tabular}[t]{@{}l@{}}- Can capture subtle semantic shifts. \\ - Density-based methods can model polysemy directly.\end{tabular} & 
\begin{tabular}[t]{@{}l@{}}- Standard embeddings conflate multiple senses into one vector. \\ - Contextual embeddings can create false positives.\end{tabular} \\ 
\midrule

\multirow{8}{*}{\begin{tabular}[c]{@{}l@{}}\textbf{Relation}\\(Metaphor/\\Metonymy)\end{tabular}} 
& \textbf{Frequency} & 
\begin{tabular}[t]{@{}l@{}}- Entropy can be a simple proxy for non-literalness.\end{tabular} & 
\begin{tabular}[t]{@{}l@{}}- Cannot distinguish figurative change from new, unrelated senses. \\ - Highly unreliable and under-explored.\end{tabular} \\ 
\cmidrule(l){2-4}
& \textbf{Topics} & 
\begin{tabular}[t]{@{}l@{}}- (Theoretical) Could potentially group figurative usages.\end{tabular} & 
\begin{tabular}[t]{@{}l@{}}- Largely unexplored for diachronic characterization. \\ - Unclear how to separate from other types of change.\end{tabular} \\ 
\cmidrule(l){2-4}
& \textbf{Graphs} & 
\begin{tabular}[t]{@{}l@{}}- Natural fit for modeling relationships between concepts. \\ - Could leverage knowledge graphs (e.g., WordNet).\end{tabular} & 
\begin{tabular}[t]{@{}l@{}}- Severely under-investigated for this task. \\ - Requires rich structured relational data, which is rare.\end{tabular} \\ 
\cmidrule(l){2-4}
& \textbf{Embeddings} & 
\begin{tabular}[t]{@{}l@{}}- Can be fine-tuned on annotated data to detect metaphors.\end{tabular} & 
\begin{tabular}[t]{@{}l@{}}- Requires labeled data for supervision. \\ - Hard to interpret *why* a vector represents a figurative use.\end{tabular} \\ 
\midrule

\multirow{8}{*}{\begin{tabular}[c]{@{}l@{}}\textbf{Orientation}\\(Amelioration/\\Pejoration)\end{tabular}} 
& \textbf{Frequency} & 
\begin{tabular}[t]{@{}l@{}}- Simple to implement via co-occurrence with seed words.\end{tabular} & 
\begin{tabular}[t]{@{}l@{}}- Easily misled by negation, irony, and sarcasm. \\ - Relies on the stability of seed words over time.\end{tabular} \\ 
\cmidrule(l){2-4}
& \textbf{Topics} & 
\begin{tabular}[t]{@{}l@{}}- (Theoretical) Could identify emerging sentiment-laden topics.\end{tabular} & 
\begin{tabular}[t]{@{}l@{}}- Not used directly for characterization; very little research. \\ - Lacks fine-grained, word-level control.\end{tabular} \\ 
\cmidrule(l){2-4}
& \textbf{Graphs} & 
\begin{tabular}[t]{@{}l@{}}- Can propagate sentiment from seed words through relations.\end{tabular} & 
\begin{tabular}[t]{@{}l@{}}- Performance is highly dependent on graph quality. \\ - Still relies on potentially unstable seed words.\end{tabular} \\ 
\cmidrule(l){2-4}
& \textbf{Embeddings} & 
\begin{tabular}[t]{@{}l@{}}- State-of-the-art for most sentiment tasks. \\ - Can measure orientation by projection onto a sentiment axis.\end{tabular} & 
\begin{tabular}[t]{@{}l@{}}- Can inherit and amplify biases from training data. \\ - Still often relies on the assumption of stable seed words.\end{tabular} \\ 
\bottomrule
\end{tabular}%
}
\end{table}

Although computational linguistics has been used to characterize \ac{lsc} for over a decade, there is still no standardized formalization of these changes. Each author proposes their own solution, and many interpretations of change are presented in a textual or semi-formal format. To contribute to this discussion, we propose a set of formal descriptions of change characterization in the next section.

\section{Formalization}
\label{sec:formalism}
Formalizing a problem is a foundational step in computer science, enabling computational reasoning and the design of algorithms. A shared formalization ensures that researchers operate with consistent assumptions and terminology, facilitating meaningful comparison across methodologies. In our review, we observed that although many studies conceptually addressed the same problem, the absence of a unified formal framework hindered direct comparison. To bridge this gap, we propose a formalization grounded in the surveyed literature and informed by linguistic theory, focusing specifically on semasiological comparative analysis; that is, comparing word meanings across time without invoking external socio-cultural knowledge.

To the core of this formalization, a central challenge arises in defining what "meaning" is. While linguistic literature often invokes terms like ‘concepts’ and ‘sentiments’ to describe meaning \citep{koch2016meaning, traugott2017semantic, Blank2003Polysemy}, these notions are rarely operationalized in a way suitable for formal modeling. This lack of clarity makes the computational implementation of semantic analysis particularly complex.

To address this, we adopt a minimal form of sense objectivism; the idea that word senses can be treated as identifiable units at specific points in time. Where the senses were conveyed by a community of speakers during the production of the corpus under study. While absolute objectivity is philosophically untenable \citep[pg.~182, pg.~32]{Geeraerts1997DiachronicPS, wittgenstein2009philosophical}, this abstraction is both methodologically necessary and practically effective (as we target static corpora). It mirrors other linguistic constructs like phonemes: not directly observable or in practical terms precise, yet essential for systematic analysis. Indeed, the very concept of semantic change presupposes a stable referent; without identifiable senses, we cannot speak coherently of change; change compared to what?

This pragmatic stance aligns with both historical linguistic practice (e.g., stating that "gay" once meant “happy” and now means “homosexual”) and computational models, which require discrete sense representations (e.g., embeddings, clusters) across time slices. Our position is not ontological but operational: sense objectivism enables formal articulation, empirical comparison, and computational modeling of lexical semantic change. Without this assumption, both theory and method lose coherence.

We build our formalization by posing the meaning of a word as a set of senses, i.e., glosses, inspired by \citet[p.~194]{HockJoseph2019Semantic}, and define all constructions in set theory. First, we assume that for a set of text corpora $T$, there is a universe of senses $S_T$ that holds all senses for these corpora. Given an alphabet $\alpha$, we define $\mathcal{S}$ as a function that takes a string from the Kleene closure of $\alpha$, where $\alpha^*=V$ (the possible vocabulary), and returns a subset of $S_T$. For instance, given a word $w \in V$ and a corpus $t \in T$, the subset of senses $\mathcal{S}(w,t)$ for the universe of senses ($S_t$) of the word $w$ in this corpus is:
\begin{equation}
\begin{split}
    \mathcal{S}:V \times T \rightarrow \wp(S_t),  \\
    \mathcal{S}(w,t) = \{s_1,s_2,...,s_k\}.
\end{split}
\end{equation}
We define that two sets of senses are different if, and only if, the symmetrical difference between a set of senses for the word $a$ in a corpus $x$ and the word $b$ in a corpus $y$ is non-empty, where $a,b \in V$ and $x,y \in T$ 
\begin{equation}
\begin{split}
    \mathcal{S}(a,x) &\neq \mathcal{S}(b,y)
    \iff \\ (\mathcal{S}(a,x) - \mathcal{S}(b,y) \neq \emptyset) &\lor (\mathcal{S}(b,y) - \mathcal{S}(a,x) \neq \emptyset),
\end{split}
\end{equation}
in simple terms, there are objective senses of the word `a' in corpus `x' that do not appear in corpus `y' for the word `b,' or vice versa. This underlying assumption is made implicitly in most previous works, even though it is rarely stated explicitly, e.g., when \cite{Hamilton2016DiachronicWE} and \cite{Inoue2022InfiniteSA} state that the word `record' has a different meaning because it gained the sense of `tape/music' when comparing the word in corpora before and after 1920.

\subsection{Semantic Change}
\label{sec:semchange}
\begin{figure}[ht]
    \centering
    \includegraphics[width=\columnwidth]{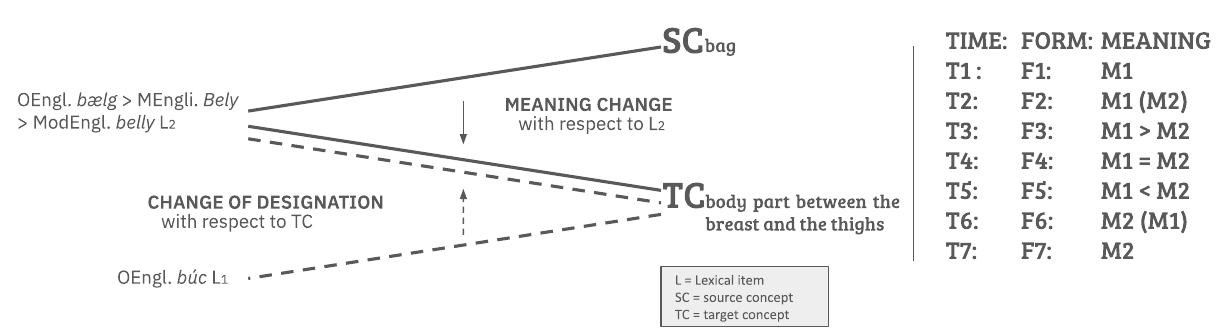}
    \vspace{.5em}
    \caption{Definition of semantic change adapted from \citet[p.~23,25]{koch2016meaning}.}
    \label{fig:koch_meaning}
\end{figure}

\citet[p.~23]{koch2016meaning} presents a definition of meaning change that combines semasiological and onomasiological perspectives. In \Cref{fig:koch_meaning}, `belly' was initially associated with the source concept `bag' (M1). Later, it acquired the meaning of the target concept `body part between the breast and the thighs' (M2). In computational linguistics, we primarily focus on the semasiological view: our interest lies in the word itself and the evolution of its meanings, rather than the origins of the concepts it represents.

The process of meaning change is depicted on the right side of the figure. At time T1, the word "belly" possessed only the meaning M1. Subsequently, at T2, it began to be used with the new meaning M2 as well. This usage of M2 increased over time (T3-T5) until it eventually became the dominant meaning (T6). Finally, at T7, the original meaning M1 was completely lost.

To formalize this definition, we say that a word $w$ had a semantic change ($\change$), if the set of senses, given by the function $\mathcal{S}$, evaluated in the corpus $t_1$ is different from the set of senses in $t_2$, given that $t_1\neq t_2$,
\begin{equation}
\change(w, t_1, t_2) = 
    \begin{cases} 
      \text{True} & \mathcal{S}(w,t_1) \neq \mathcal{S}(w,t_2)  \\
      \text{False} & \text{otherwise.} 
   \end{cases}
\end{equation}
The subset of words in vocabulary $\bar{W} \subset V$ that suffered a semantic change between corpus $t_1$ and $t_2$ is given by,
\begin{equation}
    \bar{W}_{t_1,t_2} = \{w \mid \change(w, t_1, t_2), \hspace{1em} \forall w \in V\}.
\end{equation}

These changes in lexical meaning between corpus are caused by many different phenomena, like time period (1930 and 2020, \textit{gay}), culture (Brazil and Portugal, \textit{rapariga}), and domain (engineering and biology, \textit{cell}).

\subsection{Change in Dimension}
\label{sec:dimension}
We define that a word $w$ had a changed in dimension if the set of meanings increased or decreased, with respect to the cardinality of the set, when comparing two corpus $t_1,t_2$. It means that the word suffered a broadening or a narrowing change between these corpora, i.e.,
\begin{equation}
\label{eq:dimension}
    |\mathcal{S}(w,t_1)| \neq |\mathcal{S}(w,t_2)|, \hspace{1em} w \in V.
\end{equation}
Consequentially, if $|\mathcal{S}(w,t_1)| < |\mathcal{S}(w,t_2)|$ it suffered a broadening, and if $|\mathcal{S}(w,t_1)| > |\mathcal{S}(w,t_2)|$, it suffered a narrowing.

This definition is compatible with the concept of innovative/reductive meaning change from \citet[p.~24-26]{koch2016meaning}, in which words acquire new senses (sememes) by being used with different meanings. It differs, however, from the more general notion of broadening, where words may also lose intension (the precision of their meaning), making them vaguer.

Under this definition, phenomena like sense splitting, birth, and joining are special cases of change in dimension for a single lexical item. For instance, birth or death of a sense is considered a change from or to an empty set. Split and join are instances of broadening or narrowing with respect to a word. For example, the word `plane' gained the sense of `aeroplane' after the 19th century, while previously it was only used to describe the act of planing or a flat surface. This new sense greatly increased the dimension of the word `plane', as many correlated senses, like ``traveling in an aeroplane,'' were also added.

In this survey, we define broadening as the degree of polysemy in a word. This is aligned with the \ac{nlp} literature, given its application in data analysis and its measurable nature, compared to the notion of intension. We differ from the definition in \citet{Tang2013SemanticCC}, where broadening is a superclass of metaphorization and metonymization, as the literature reviewed here does not perform this aggregation.

\subsection{Change in Relation}
\label{sec:relation}
Metaphorization or metonymization occurs when a word takes on a new meaning that inherits qualities from its original meaning through a figurative relationship the speaker aims to convey \citep[p.~3-6]{lakoff2008metaphors}. The original meaning is invoked by establishing a material or abstract relation with the initial sense, where the listener is able to resolve the non-literal target concept \cite{Fass1988MetonymyAM}. Material relations are built on physical-world contiguity (e.g., replacing an institution with its location, as in ``Washington decided..."). In contrast, abstract relations are formed through conceptual similarity (e.g., applying human body parts to objects, as in ``the arm of the chair'').

The theory defining metaphor and metonymy is often informal, relying on subjective perception or overlapping definitions. For example, \citet{Choi2021Melbert} defines metaphor as a case where the ``literal meaning of a word is different from its contextual meaning,'' which also applies to metonymy. As discussed by \citet[p.~42-51]{koch2016meaning}, there is no precise rule to define a metonymy. Some computational work, such as \citet{Schlechtweg2017GermanIF}, measures a word's entropy to capture metaphorization, but it is not clear that this measure can differentiate a metaphor from a metonymy or a completely new, unrelated sense (e.g., `Apple' in a technology context).

Given these considerations, we define that a word $w$ has changed in relation if the aggregated sum of the relational similarity ($\mathit{l}: S \times S \rightarrow \mathbf{R}$) between its senses changes between corpora. The function $\mathit{l}$ takes two senses, $s_i$ and $s_j$, and returns a real value corresponding to their material or abstract relation (a greater value means a stronger relation). The total relation for a word is:

\begin{equation}
\label{eq:relation}
\relation(w, t)= \left\{ \begin{aligned}
  \sum^{N-1}_{i=1}\sum^{N}_{j=i+1}{\mathit{l}(s_i, s_j)} \hspace{3em} s_i,s_j \in \mathcal{S}(w, t)\\
  0 \hspace{6em} N \leq 1,
\end{aligned} \right.
\end{equation}
where $N=|\mathcal{S}(w, t)|$.

A word $w$ increases in relation between corpora $t_1$ and $t_2$ if $\relation(w,t_1) < \relation(w,t_2)$, and decreases if the opposite occurs. Since we lack methods to perfectly differentiate material and abstract relations, and since a word can undergo both metaphorization and metonymization, we simplify the interpretation: if a word's figurative usage increases, it "increases in relation"; if its figurative usage decreases, it "decreases in relation."

\subsection{Change in Orientation}
\label{sec:orientation}
A word's change in orientation (positive or negative connotation) reflects amelioration or pejoration in its usage from one corpus to another. This is measured by analyzing how the collective perception of a word's meaning evolves \cite[p.~227-229]{campbell2013historical}. A classic example is `terrific,' which originally meant `terrible' but shifted to mean `formidable' or `impressive.'

While some frameworks like \citet{Buechel2016FeelingsFT} use a multi-dimensional Valence-Arousal-Dominance setting, most sentiment analysis research focuses on a simpler polarity setting \citep{Cui2023SurveyOS}. Therefore, we restrict our formalism to this more general positive/negative view to align with current research and applications \citep{Susanto2020TheHM}.

We define that a word $w$ had changed in orientation, if for some notion of orientation $\feeling$ (feeling), there is a mapping from a word $w$ in a corpora $t$ to an orientation value, for example, in case of positive, neutral, or negative feeling,
\begin{equation}
\feeling:V \times T \rightarrow \{-1, 0, +1\},
\end{equation}
the orientation changes between corpora $t_1,t_2$,
\begin{equation}
\feeling(w, t_1) \neq \feeling(w, t_2).
\end{equation}
Analogous to the previous definition, if $\feeling(w, t_1) < \feeling(w, t_2)$ it suffered an amelioration, and a pejoration if $\feeling(w, t_1) > \feeling(w, t_2)$.

This task relies on a subjective notion of orientation captured by the function $\feeling$. Generally, studies define `orientation' as the proximity to positive or negative seed words. However, the problem is broader than just polarity, as words can be classified along other emotional dimensions.

\subsection{Semantic Change Identification}
Core to computational approaches, words in a corpus are not compared directly to one another; instead, they are first represented using one of the many methods reviewed in this survey. A common step taken by all authors is to represent the word `a` in corpus `x`, denoted as $\representation(a,x)$, and then compare this representation to $\representation(b,y)$. For example, \cite{moss2020detecting} represents the word `a` as a Gaussian embedding obtained after training a language model on corpus `x`. Similarly, \cite{Ehmller2020SenseTD} represents `a` using a co-occurrence graph derived from corpus `x`. Meanwhile, \cite{Inoue2022InfiniteSA} partitions the original corpus into 20-year intervals and, for each resulting corpus `x`, obtains a distribution of topics for the word `a`.

In this context, we now define in computational terms the task of semantic change characterization. Consider a function $f_{\change}$ that takes two representations $\representation$ (e.g., embeddings, graphs, etc.) of the word $w$ in corpora $t_1,t_2$. We define the problem of semantic change identification as assigning a value ($y$) to this word,
\begin{equation}
    f_{\change}(\representation(w,t_1), \representation(w,t_2)) \rightarrow y.
\end{equation}

\citet{Schlechtweg2020SimulatingLS} proposed two notions of semantic change: binary classification, where $y \in \{0,1\}$, and graded (continuous) semantic change, where $y \in \mathbf{R}$ is a distance between word representations\footnote{While \citet{Schlechtweg2020SimulatingLS} considered graded change for a particular purpose (i.e., Jensen-Shannon distance), we generalize $y$ to any applicable distance measure.}.

The \textbf{binary} classification task aligns with classic \ac{lsc} literature \citep[p.~113]{blank2012prinzipien} and human perception, which is often influenced by sensory thresholds \citep[p.~34]{smith2008biology}. This view is interested in the gain or loss of senses. There is no general threshold criterion for the binary setting; it's not enough for representations to be different, the magnitude of this difference should align with human judgment.

The \textbf{graded} task originates from computational linguistics, where a natural way to compare objects is by measuring the distance between their representations. This approach allows for a more in-depth analysis of change, as the grade of change can be compared across a wide range of contexts.

\subsection{Semantic Change Characterization}
Deriving from the reviewed literature, we define semantic change characterization as a classification problem where we need to assign identified semantic changes to categories, such as $f_\dimension,f_\relation,f_\orientation$,
\begin{equation}
    f_x(\representation(w,t_1), \representation(w,t_2)) \rightarrow y,\hspace{2em} x \in \{\dimension,\relation,\orientation\}.
\end{equation}

While the first equation only measured some notion of distance between the representations, for characterization we first apply the type function on the representation, to then compare. For example in \citet{Buechel2016FeelingsFT} we have a orientation characterization, e.g.,
\begin{equation}
    f(\orientation(\representation(w,t_1)), \orientation(\representation(w,t_2))) \rightarrow y,
\end{equation}
where the $\orientation$ function is the Valence-Arosal-Domance framework and they measure the distance $f$ through the $\beta$-coefficient of a linear regression.

Again we set two possibilities of change, binary and graded. In this setting the binary form allow us to answer interesting questions in an automatic manner, for example, in which point in time the word `head' started being used in a figurative manner? Another possibility is to track the precise corpus were words pejorated, and then manually understand the socio-linguistic process to this phenomena, saving many hours of historiography work.

\subsection{Framework Evaluation}
\label{subsec:framework}
\begin{figure}[ht]
    \centering
    \includegraphics[width=0.7\columnwidth]{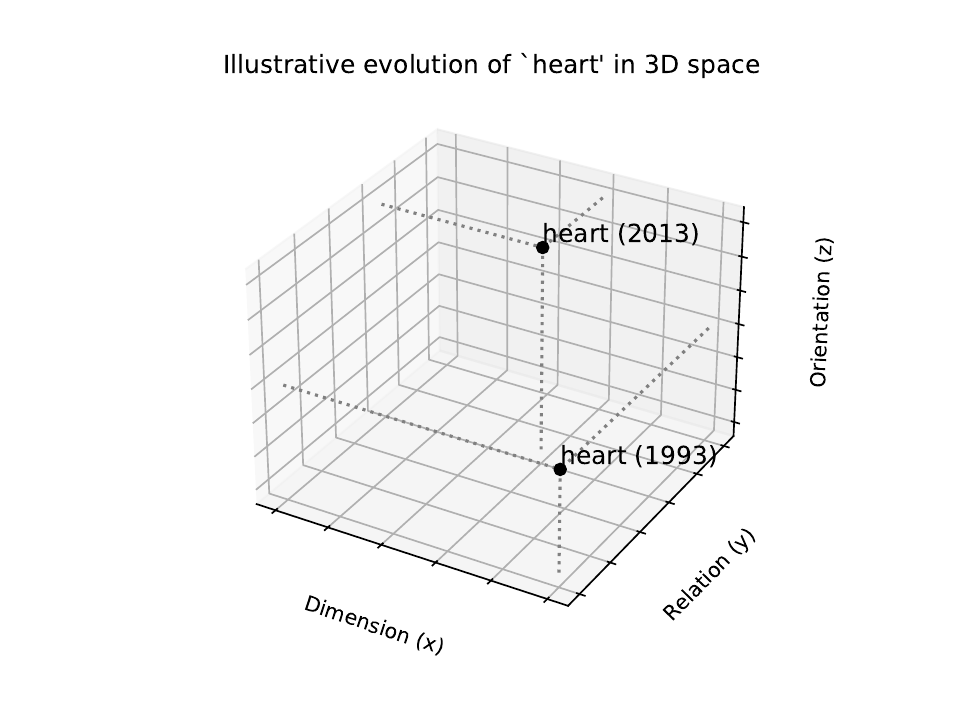}
    \caption{Illustrative example of the word `heart' changing over time. Metaphorization, Amelioration and Broadening can occur for the same word, depending on the senses it gained/lost.}
    \label{fig:semantic-evolution}
\end{figure}

The proposed framework enables a spatial understanding of a word's change and direct statistical comparison of corpora. In \Cref{fig:semantic-evolution}, we illustrate this analysis with a possible evolution of the word `heart' between two corpora from different times (1993 and 2013). A gradual analysis of this evolution can provide a detailed picture across the three poles of change.

To illustrate the framework's practical aspects, we investigate two corpora: SEMCOR \citep{Miller1993ASC} and MASC\footnote{\url{https://anc.org/data/masc/corpus/}}. SEMCOR is a sense-tagged corpus of ~200,000 words created in 1993. MASC is a manually annotated sub-corpus of ~500,000 words of modern American English released around 2013.

While the corpora were produced at different times (1993 and 2013), we do not claim this analysis is sufficient to discover all examples of historical \ac{lsc}. The corpora have narrow coverage and are inadequate for robust diachronic analysis. Additionally, WordNet doesn't cover all existing senses for a given word. We use these datasets for illustrative purposes only. For the purpose of this demonstration we define as corpus $t_1$, SEMCOR and MASC as $t_2$, with the time difference of 20 years between corpus.

SEMCOR and MASC are corpora manually annotated using WordNet senses. We obtain annotations for the poles of change from two lexical resources: ChainNet, a WordNet extension linking senses by metaphor and metonymy, and SentiWordNet, which provides a positive/negative score for every sense in WordNet. This step allows us to skip sense inference and focus on demonstrating the framework. For automatic WordNet annotation, we suggest \cite{Navigli2017WordSD}. For work on annotating data for the poles of change independently of a lexical resource, we suggest \cite{Sa2024SemanticCC}. From SEMCOR and MASC, we computed occurrences of `heart' and `plane,' two well-known words that have changed semantically.

\begin{table}[ht]
\centering
  \caption{Distribution of WordNet senses for the word `heart' in SEMCOR and MASC. The score column indicates positive score for that sense from SentiWordNet, figurative column indicates if the meaning is figurative or literal.}
 {
    \begin{tabular}{lcccc}
    \hline
    Sense & Score (+) & Figurative & SEMCOR & MASC \\\hline
heart.n.03    & 0.25 & False & 0.121212 & 0.901408\\\hdashline
kernel.n.03   & 0.25 &  True & 0.030303 & 0.070423\\\hdashline
center.n.01   & 0.00 &  True & 0.060606 & 0.000000\\\hdashline
heart.n.10    & 0.00 &  True & 0.000000 & 0.028169\\\hdashline
heart.n.01    & 0.00 &  True & 0.439394 & 0.000000\\\hdashline
heart.n.02    & 0.00 & False & 0.348485 & 0.000000\\
    \hline
    \end{tabular}}
  \label{tab:heart}
\end{table}

According to SentiWordNet, `heart' has two positive senses: courage (heart.n.03) and the vital part of an idea (kernel.n.03). From ChainNet, the absolute literal meanings are the organ (heart.n.02) and courage (heart.n.03). We can observe that in MASC, the literal meaning `organ' (heart.n.02) has no occurrences, while figurative usage becomes dominant. The word also gained a new sense (heart.n.10, the playing card suit).

From \Cref{tab:heart}, we can compute the equations to verify the semantic change across the corpus. We first start with the dimension test. From \Cref{eq:dimension} we have:
\begin{equation}
|\mathcal{S}(w,t_1)| \stackrel{?}{=} |\mathcal{S}(w,t_2)|
\end{equation}

\begin{equation}
|\mathcal{S}(\text{heart},\text{SEMCOR})| \stackrel{?}{=} |\mathcal{S}(\text{heart},\text{MASC})|
\end{equation}

\begin{equation}
\begin{gathered} 
|\{\texttt{heart.n.03}, \texttt{kernel.n.03}, \texttt{center.n.01}, \texttt{heart.n.01}, \texttt{heart.n.02}\}| \\
\stackrel{?}{=} \\
|\{\texttt{heart.n.03}, \texttt{kernel.n.03}, \texttt{heart.n.10}\}|
\end{gathered}
\end{equation}

\begin{equation*}
5 > 3, \text{we conclude that it narrowed.}
\end{equation*}

To evaluate changes in Orientation, we define the feeling function $\feeling$ as the weighted sum of the positive scores of a word's senses. Specifically, we compute $\feeling$ as the sum of $\text{positive}(s_i) \cdot p(s_i)$, where $\text{positive}(s_i)$ is the score of sense $s_i$, and $p(s_i)$ is its associated probability. A sense is considered positive if its score is greater than 0.

We use the weighted sum as a proxy for the prototypical sense, assuming that more frequently used senses have a greater influence on the perceived feeling of the word.

\begin{equation}
\feeling(w, t) = \frac{1}{N}\sum_{i=1}^N p(s_i) \text{positive}(s_i), \hspace{1em}s_i \in \mathcal{S}(w, t)
\end{equation}

Plugging the values into our equations, we have:
\begin{equation}
\feeling(\text{heart}, \text{SEMCOR}) \stackrel{?}{=} \feeling(\text{heart}, \text{MASC})
\end{equation}

\begin{equation}
\begin{gathered} 
1 \times 0.12 + 1 \times  0.03 + 0 \times 0.06 + 0 \times 0.0 + 0 \times .44 + 0 \times 0.35\\
\stackrel{?}{=} \\
1 \times 0.9 + 1 \times  .07 + 0 \times 0.0 + 0 \times 0.02 + 0 \times 0.0 + 0 \times 0.0
\end{gathered}
\end{equation}

\begin{equation*}
    0.15 < 0.97, \text{we observe an amelioration.}
\end{equation*}

\begin{figure}[ht]
    \centering
    \includegraphics[width=\columnwidth]{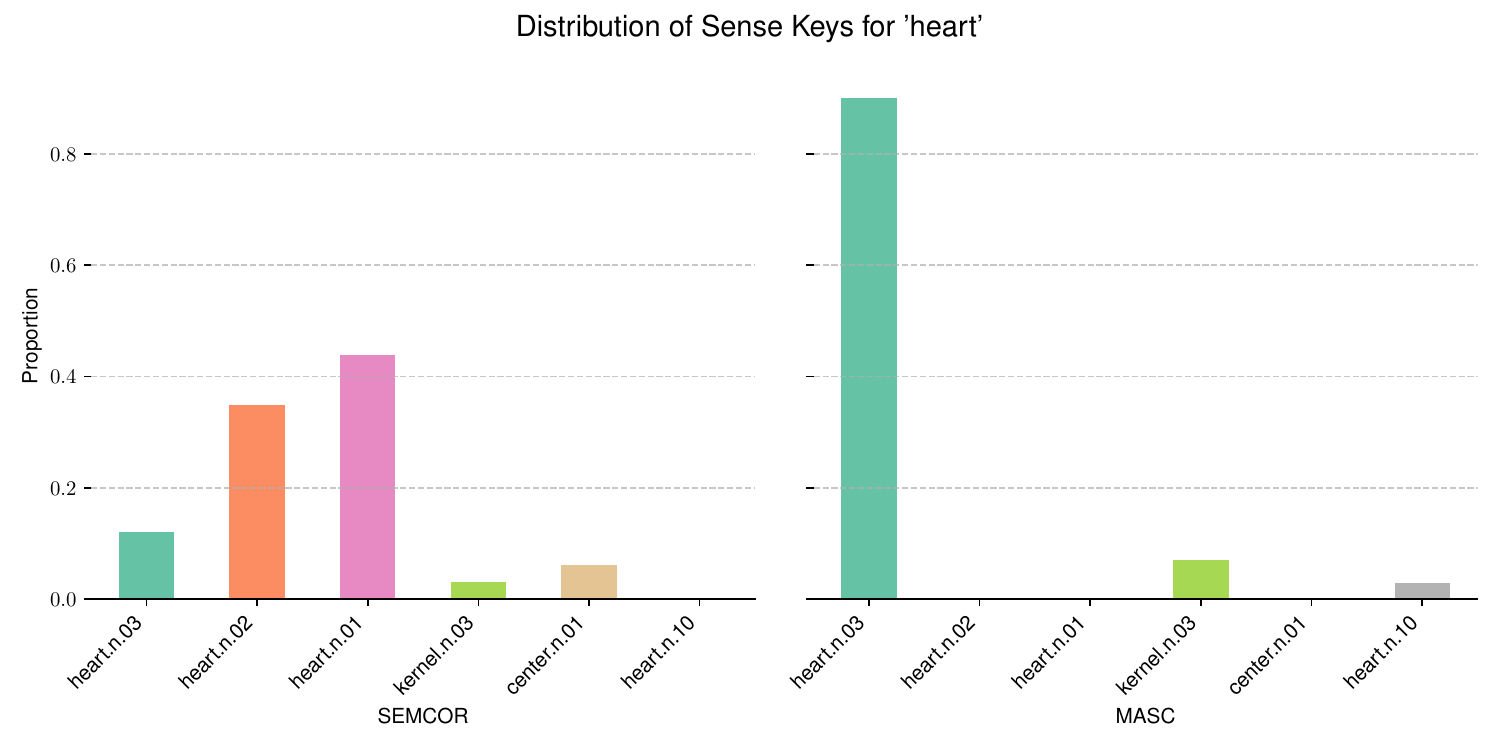}
    \vspace{1em}
    \caption{Change in meaning distribution for the word `heart' in SEMCOR and MASC.}
    \label{fig:heart}
\end{figure}

\Cref{fig:heart} illustrates the change in meaning distribution for `heart.' Under our framework, the word becomes more metaphorical (increasing in Relation, $\relation$), narrowed for these corpora (decreasing its Dimension, $\dimension$), and more positive (increasing in Orientation, $\orientation$).

\begin{table}[ht]
\centering
  \caption{Distribution of WordNet senses for the word `plane' in SEMCOR and MASC. The score column indicates positive score for that sense from SentiWordNet, figurative column indicates if the meaning is figurative or literal.}
 {
    \begin{tabular}{lcccc}
    \hline
    Sense & Score (+) & Figurative & SEMCOR & MASC \\\hline
plane.n.03    & 0.000 & True  & 0.046512 & 0.000000\\\hdashline
airplane.n.01 & 0.000 & False & 0.488372 & 0.909091\\\hdashline
flat.s.01     & 0.375 & False & 0.046512 & 0.000000\\\hdashline
plane.v.01    & 0.000 & False & 0.046512 & 0.000000\\\hdashline
plane.n.02    & 0.000 & False & 0.372093 & 0.090909\\
    \hline
    \end{tabular}}
  \label{tab:plane}
\end{table}

For `plane,' we note that while SEMCOR reports five distinct senses, MASC presents only two. In MASC, `plane' has lost its usage as a surface (flat.s.01), a level of existence (plane.n.03), and a type of cut (plane.v.01), while the prototypical usage of `airplane' has become more prevalent.

For the Relation dimension, we compute $l$ by counting the number of linked senses as described in \Cref{eq:relation}.

\begin{equation}
    \relation(\text{plane}, \text{SEMCOR}) \stackrel{?}{=} \relation(\text{plane}, \text{MASC})
\end{equation}

\begin{equation}
\begin{gathered}
\relation(\text{plane}, \text{SEMCOR}) = \mathit{l}(\text{plane.n.03}, \text{plane.n.02}) = 1\\
\stackrel{?}{=} \\
\relation(\text{plane}, \text{MASC}) = 0
\end{gathered}
\end{equation}

\begin{equation*}
   1 > 0, \text{so the sense is less metaphorical and decreased in relation.}
\end{equation*}

\begin{figure}[ht]
    \centering
    \includegraphics[width=\columnwidth]{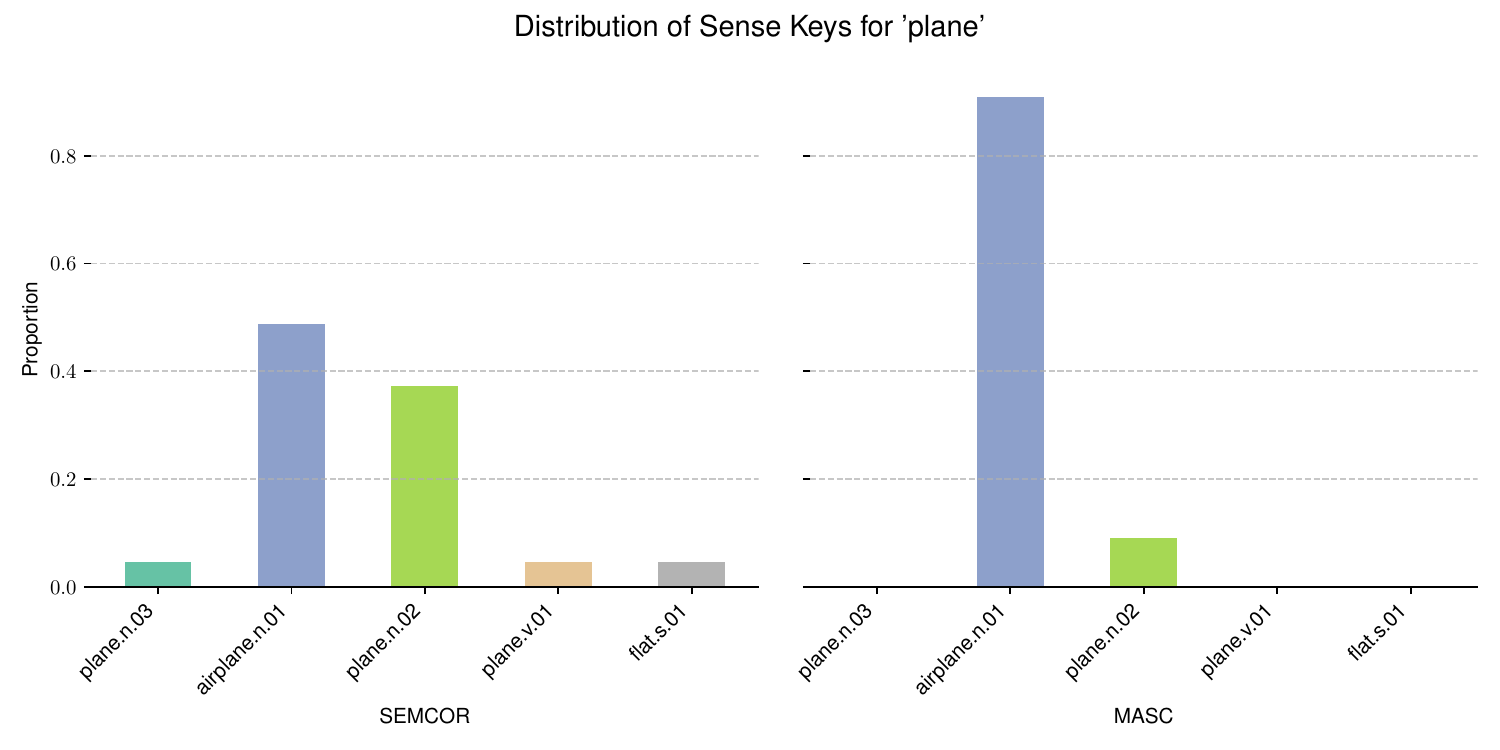}
    \vspace{1em}
    \caption{Change in meaning distribution for the word `plane' in SEMCOR and MASC.}
    \label{fig:plane}
\end{figure}

In the \Cref{fig:plane}, we can see the narrowing effect this word suffered for this corpus. The decreased usage as a positive meaning (flat.s.01) causes a decrease in its Orientation ($\orientation$).

\section{Discussion}
\label{sec:discussion}
In this section, we synthesize the reviewed approaches for change characterization and discuss their limitations according to the dimension, relation, and orientation aspects we have introduced and formalized.

\subsection{Dimension}
The majority of studies found in this survey propose methods to characterize how words change in dimension, while the other poles are less studied. Change in dimension seems easier to detect because it doesn't rely on subjective interpretation, such as relation or feeling, but mainly on a countable quantity (see \Cref{sec:formalism}). Work along this axis often relies on word sense induction, a research area with well-developed tooling and many data sources, such as dictionaries and synsets, used for supervised learning.

The key factor differentiating the approaches is the word representation method. We identified four main types—frequency, topics, graphs, and embeddings—each with different trade-offs in implementation cost, representation power, and interpretability. It is difficult to compare the effectiveness of each method for the current state of the art, given the qualitative nature of reported results, the use of different corpora, and the need for external knowledge. However, our analysis reveals that frequency- and embedding-based methods appear to produce better results for identifying neologisms (new words), while topic- and graph-based approaches are more effective at detecting sememes (new senses) over time.

A limitation for approaches based on dictionaries is the temporal delay between a neologism's first appearance and its inclusion in dictionaries. Another limitation is the presence of metaphorical senses at the same level as literal ones, increasing the complexity of distinguishing between 'dimension' and 'relation.' Some authors prefer to group these two poles into one \citep{Tang2013SemanticCC}. For topic- and graph-based approaches, the limitation is that interpreting the meaning of a topic or a graph branch is not always clear and can require expert intervention.

\subsection{Relation}
Outside of this \ac{lsc} survey, a vast body of work proposes methods to predict if a word in a sentence has a metaphorical sense \citep{Heintz2013AutomaticEO, Shutova2013UnsupervisedMI, Hovy2013IdentifyingMW}. These approaches analyze texts but don't consider temporal information, so they cannot determine if a word has undergone metaphorization over time. Some works claim to detect the metaphorization of words (e.g., \citep{Tredici2016TracingMI}), but they rely on the general hypothesis that any sense change implies a metaphorical change, which still needs to be demonstrated.

For metonymization, we could not find studies that performed metonymy identification in a comparative setting. However, related recent studies investigate metonymy usage with various methods for material relation identification \citep{Nastase2012LocalAG, BaldiniSoares2019MatchingTB, AlexMathews2021ImpactOT}. We expect that future investigations can extend these approaches to detect word metonymization.

Among the four meaning representation methods, frequency-based approaches were the first investigated for relation characterization. However, their underlying hypothesis doesn't differentiate between new metaphorical/metonymical usage and new homonymous usage, making interpretation difficult. More recent works employed embeddings, but interpreting vectors in high-dimensional space remains a challenge. Finally, graphs, well-known for representing relations, remain under-investigated for identifying changes in relations.

\cite{Tang2013SemanticCC} mainly rely on dictionaries of heuristics to infer metaphoricity, which served for qualitative analysis, but still require more extensive validation. A single work used ground-truth annotations for metaphors \citep{Schlechtweg2017GermanIF}, but did not reported a final precision of the proposed methodology. 

As noticed, studies on the 'relation pole' are few and inconclusive. The research community's limited interest in this topic could be justified by three reasons: (I) the difficulty of formally defining metaphor and metonymy from a computational perspective, (II) the rarity and insufficiency of annotated corpora for relation detection, and (III) the nonexistence of standard metrics to measure and compare approaches.

\subsection{Orientation}
Interest in orientation analysis is increasing with the wide adoption of social networks. Orientation is usually determined using a feeling function ($\feeling$) that relies on seed words and the assumption that these seed words don't vary over time. As we illustrated, word meanings evolve and can differ between communities. Most works found in this survey addressed the change in orientation for particular words \cite{Xiao2023HaveTC, MillerLewis2021WordsDF}.

The example of `sick` shows that a word can have two opposite orientations depending on the context. Creating corpora that include all these variations with statistical significance is a complex task. The annotation process can suffer from many biases, both from annotators and data. Selected seed words may not generalize across all contexts or long time spans, and a series of misleading results can occur, as presented in \citet{Antoniak2021BadSE}.

Of the four meaning representation methods, embedding- and frequency-based approaches are the most common. Few works use graphs, and we could not find any approach that uses topic analysis for detecting word orientation. The preference for embeddings and frequency is likely because current approaches compute distances or co-occurrences with seed words, which is easier with these methods.

While there is a huge body of work on sentence-level sentiment analysis, fewer studies infer sentiment at the word level \citep{Guo2021AnOO}, and only a small fraction of those do so in a comparative manner, as we have assessed in this survey. While works in topics provide lexical entries that could be used similar to what was done in embeddings, this was not investigated so far.

An important outcome of this survey is that there is some harmonization in how change in feeling is defined and applied. However, we did not observe a harmonized approach to error analysis for the feeling function. Some works adopt a binary orientation function (positive, negative), while others use multidimensional orientations (valence, arousal, dominance) \citep{Buechel2016FeelingsFT}. Further study is needed to analyze which function ($\feeling$) is most suitable for characterizing change.
\\

A converging point from the many approaches analyzed is the difficulty of evaluating and comparing results because, for most tasks, no gold standard exists. Many papers have raised this problem \citep{Schlechtweg2017GermanIF}, but some initiatives to create reference datasets are underway. Another point is multilingualism; the vast majority of experiments use only English documents, making it difficult to generalize methods to other languages. The recent emergence of new-generation language models has opened up new possibilities for identifying and characterizing semantic changes, but this area of research is still under-explored.

We begin this survey with five objectives, which we believe have been archived:
\begin{enumerate}
    \item We discussed the limitations of current representations and methodologies in this section.
    \item We evaluated the strengths of various approaches and the complexity of their computational interpretation in \Cref{tab:pros_cons}.
    \item We proposed formal definitions in \Cref{sec:formalism} to reduce ambiguity.
    \item We demonstrate our proposed formalization with real data in \Cref{subsec:framework}.
\end{enumerate}

\section{Conclusion}
\label{sec:conclusion}
In this survey, we reviewed current work on semantic change characterization with respect to the mainstream typology (broadening/narrowing, amelioration/pejoration, and metaphorization/metonymization). We observed four major approaches to representing meaning—frequency, topics, embeddings, and graphs—and their respective strengths and drawbacks.

After analyzing the literature, we observed that many tasks are under-explored or addressed without systematic evaluation. Therefore, we introduced a formal definition for the task of semantic change characterization, a more complex instance of the semantic change identification problem. This new task is extremely relevant for the field of \ac{lsc}, as this study and other surveys have argued.

The suggested tasks are crucial for both linguistic change research and \ac{nlp}. Numerous applications, such as sentiment analysis (orientation), semantic similarity (relation), and word sense inference (dimension), call for a deeper understanding of these representational characteristics. Additionally, awareness of how these changes affect word representations leads to better knowledge and interpretability of these systems. For word embeddings specifically, a better understanding of how these changes affect vector space geometries could improve our understanding of the semantics of vectors.

As future work, we expect to investigate the relation pole to better differentiate categories of change and further explore how the four representational methods relate to the three poles. Another important goal is the consolidation of datasets and metrics by which semantic change characterization can be compared across methodologies. In this survey, we performed an initial investigation of existing sense-annotated corpora; in future studies, we expect to analyze words in more comprehensive corpora.

Finally, this survey has limitations in its coverage of \ac{lsc} and computational linguistics. Since the study of word meaning involves multiple, rapidly evolving fields, we focused on key works related to computational characterization and traditional meaning representations. The rise of new-generation language models and new theories of language development offers further opportunities for studying semantic change, but this area is still not fully explored.

\section*{Acknowledgments}
The research reported in this publication was supported by the Luxembourg National Research Fund (FNR), project D4H grant number PRIDE21/16758026

\section*{Use of AI tools}
The production of this manuscript used AI tools solely to improve grammar and readability, as English is not the authors’ native language.

\bibliographystyle{nlelike}
\bibliography{NLEguide}

\label{lastpage}

\end{document}